\documentclass[10pt,twocolumn,letterpaper]{article}

\usepackage[pagenumbers]{wacv} 

\usepackage{amsmath}
\usepackage{amssymb}
\usepackage{pifont}

\newcommand{\cmark}{\textcolor{green!60!black}{\ding{51}}}
\newcommand{\xmark}{\textcolor{red!70!black}{\ding{55}}}

\usepackage{tikz}
\usetikzlibrary{positioning}
\usepackage{pgfplots}
\pgfplotsset{compat=1.18}
\usepgfplotslibrary{groupplots}
\usetikzlibrary{arrows.meta}
\usepackage{multirow}
\usepackage{booktabs}
\usepackage{colortbl}
\usepackage{makecell}
\usepackage{graphicx}
\usepackage{float}

\newcommand{\methodrule}{\specialrule{0.25pt}{0.4pt}{0.4pt}}

\usepackage{algorithm}
\usepackage{algpseudocode}
\usepackage{listings}
\lstdefinestyle{gode}{
  basicstyle=\ttfamily\scriptsize,
  keywordstyle=\color{blue!60!black}\bfseries,
  commentstyle=\color{green!45!black}\itshape,
  stringstyle=\color{red!55!black},
  numbers=left, numberstyle=\tiny\color{gray}, numbersep=6pt,
  breaklines=true, columns=fullflexible, showstringspaces=false,
  language=Python, frame=tb, framesep=4pt, aboveskip=6pt, belowskip=4pt,
  morekeywords={with,as,assert,lambda,yield}
}
\definecolor{wacvblue}{rgb}{0.21,0.49,0.74}
\usepackage[pagebackref,breaklinks,colorlinks,allcolors=wacvblue]{hyperref}

\def\wacvPaperID{804} 
\def\confName{WACV}
\def\confYear{2027}

\title{GoDe: Gaussians on Demand for Backbone-Agnostic Progressive Compression}

\author{
Francesco Di Sario\\
Universit\`a degli Studi\\
di Torino, Italy\\
{\tt\small francesco.disario@unito.it}
\and
Riccardo Renzulli\\
Universit\`a degli Studi\\
di Torino, Italy\\
{\tt\small riccardo.renzulli@unito.it}
\and
Marco Grangetto\\
Universit\`a degli Studi\\
di Torino, Italy\\
{\tt\small marco.grangetto@unito.it}
\and
Akihiro Sugimoto\\
National Institute of Informatics\\
Tokyo, Japan\\
{\tt\small sugimoto@nii.ac.jp}
\and
Enzo Tartaglione\\
LTCI, T\'el\'ecom Paris\\
Institut Polytechnique de Paris, France\\
{\tt\small enzo.tartaglione@telecom-paris.fr}
}

\begin{document}
\maketitle\begin{abstract}
Splat-based radiance fields have rapidly diversified in primitive type,
kernel support, and appearance parameterization.
This diversity makes compression difficult: most codecs are single-rate,
requiring a separately optimized model for each target bitrate, while
progressive codecs remain tied to representation-specific assumptions
that do not transfer across backbones.
We introduce GoDe (Gaussians on Demand), a progressive compression
framework whose central contribution is \textit{backbone agnosticism}: it
reorganizes any trained differentiable splat representation into a single
scalable bitstream without modifying the backbone or re-deriving the
compression structure for each new primitive type.
GoDe builds a nested level-of-detail hierarchy via masked-gradient
importance ranking, refines all levels with one quantization-aware
fine-tuning stage, and stores the base layer and enhancement layers as
progressive chunks compressed with shared entropy dictionaries.
Any decoded prefix is directly renderable, so one training pipeline
provides multiple rate--distortion operating points without per-rate
retraining.
Across five heterogeneous backbones and three standard benchmarks, GoDe achieves competitive rate–distortion performance while enabling faster rendering and substantially faster progressive decoding than existing methods. 
\noindent\textbf{Project page:}
\texttt{https://gaussians-on-demand.github.io/}
\end{abstract}
    
\section{Introduction}
\label{sec:intro}

\begin{figure}[t]
    \centering
    \includegraphics[width=\linewidth]{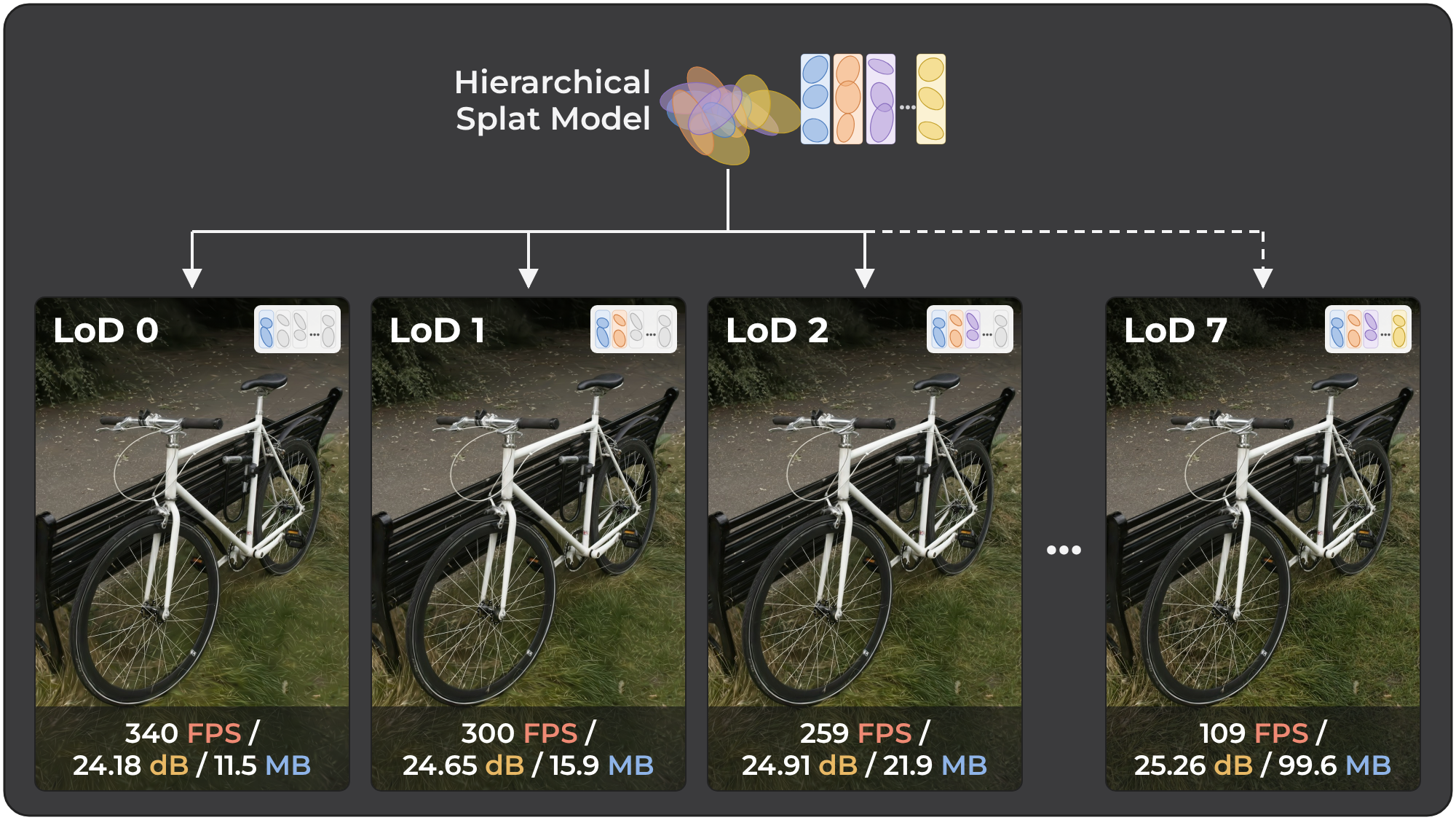}
    \caption{\textbf{One training run, a full rate curve.} GoDe constructs
    a progressive hierarchy from any trained splat representation; each
    decoded prefix yields a distinct quality--size--speed operating point
    with no per-rate retraining.}
    \label{fig:teaser}
\end{figure}

Novel view synthesis has been transformed by Neural Radiance
Fields~\cite{mildenhall2020nerf} and, more recently, by 3D Gaussian
Splatting~\cite{kerbl20233d}.
Splat-based radiance fields have since diversified rapidly beyond the
original Gaussian formulation, with methods differing in primitive type~\cite{held2026triangle, govindarajan2025radiant, scaffoldgs},
kernel support~\cite{liu2025deformable}, and view-dependent appearance model~\cite{di2026spherical}.
This diversity improves reconstruction and rendering quality, but it also fragments compression: a codec designed around one set of internal parameters is difficult to transfer to another backbone.
Compression is necessary for deployment, since uncompressed scenes can
occupy hundreds of megabytes.
Most codecs are single-rate~\cite{3DGSzip2024}: each memory budget
requires its own optimized model.
This differs from progressive rendering and level-of-detail
methods~\cite{shi2024lapisgs, cheng2025clod}, which reduce the number of
rendered primitives at runtime but do not define a progressively
decodable compressed representation.
For streaming and adaptive storage, the useful object is a bitstream
whose prefixes are valid renderable models.
Progressive compression provides exactly this structure: a base layer can
be decoded first, and enhancement layers refine the same scene
incrementally.
Existing progressive codecs~\cite{tseng2026csgaussian, chen2026pcgs} achieve strong
rate--distortion results within the Scaffold-GS family, but their entropy context is conditioned on internal anchor feature vectors that have no analog in other primitive types.
As new splat representations continue to appear, tying progressive compression to one specific backbone reintroduces the same fragmentation that motivated compression in the first place. The missing ingredient is a progressive compression structure that operates on primitives as atomic units, without requiring access to their internal parameterization. In this work, we introduce GoDe (Gaussians on Demand), a progressive
compression framework for differentiable splat representations
(Fig.~\ref{fig:teaser}).
The key insight is that three operations are available for any
backbone that supports differentiable rendering: gradient-based
importance ranking, quantization-aware optimization, and serialization
of primitive parameters.
GoDe composes these into a single pipeline that produces a progressive
bitstream from any trained model.
Decoding a level requires only decompressing and concatenating the chunks up to that level: no optimization is performed at decode time.
To apply GoDe to a new backbone, one specifies the primitive grouping
and the quantization rule for each parameter group. GoDe composes gradient-based importance ranking, quantization-aware
fine-tuning, and entropy coding into the first progressive compression
framework that transfers across heterogeneous splat backbones.
The critical design choice is operating at the level of primitives as
atomic units: by requiring only differentiable rendering, parameter serialization, and a lightweight
per-backbone specification of parameter groups, GoDe avoids the representation-specific assumptions that
limit existing progressive codecs to a single backbone family.
One training pipeline produces a full rate--distortion curve; one
compressed file exposes multiple operating points at deployment time. Across five backbones, GoDe provides competitive rate--distortion curves,
fast rendering on explicit representations, and fast progressive decoding.
To summarize, our contributions are:
\begin{itemize}
    \item \textbf{A backbone-agnostic progressive compression framework}
    for differentiable splat representations, producing multiple decoded
    prefixes from a single training pipeline without modifying the
    underlying backbone or re-deriving the compression structure per
    primitive type.
    \item \textbf{A simple, transferable pipeline} composed of
    gradient-informed importance ranking, stochastic quantization-aware
    fine-tuning over nested levels, and chunk-based entropy coding with
    shared dictionaries.
    \item \textbf{Evaluation across five heterogeneous backbones},
    demonstrating competitive rate--distortion, high rendering speed, and decoding times one to two
    orders of magnitude faster than existing methods.
\end{itemize}

\section{Related Work}
\label{sec:related}

\noindent\textbf{Splat-based radiance fields.}
3D Gaussian Splatting~\cite{kerbl20233d} achieves real-time novel view
synthesis via differentiable rasterization of anisotropic Gaussian
primitives.
Its success has led to a diverse family of splat representations.
Scaffold-GS~\cite{scaffoldgs} generates Gaussians from neural
anchors, Octree-GS~\cite{ren2024octree} adds a hierarchical anchor
organization, and Triangle Splatting~\cite{held2026triangle} replaces
ellipsoidal kernels with flat triangle primitives.
Other works change the local support or the optimization process:
Deformable Beta Splatting~\cite{liu2025deformable} uses bounded
shape-adaptive Beta kernels, while MCMC-based
training~\cite{kheradmand2024d} replaces the standard densification
heuristics with a Markov chain sampling strategy for primitive
placement.
Appearance is also representation dependent: classical 3DGS stores
spherical harmonics per primitive~\cite{fridovich2022plenoxels}, but alternative representations have
been explored, ranging from neural feature codes decoded by a small
MLP~\cite{scaffoldgs}, to Spherical Voronoi
tessellations~\cite{di2026spherical}.
These choices are not interchangeable at the parameter level, which is
why compression methods that depend on a particular internal structure
are difficult to reuse across backbones.

\noindent\textbf{Progressive rendering and level of detail.}
PRoGS~\cite{zoomers2025progs} prioritizes splats by contribution for progressive
streaming.
LapisGS~\cite{shi2024lapisgs} constructs a layered representation for adaptive
bandwidth rendering.
CLoD-GS~\cite{cheng2025clod} learns distance-dependent opacity decay for smooth
quality scaling from a single model.
Octree-GS~\cite{ren2024octree} and
Hierarchical-GS~\cite{hierarchicalgaussians24} dynamically select the
level of detail at render time based on viewing distance.
These methods adapt which primitives are rendered or at which resolution
a scene is accessed, but they do not define a compressed bitstream in
which each prefix is a valid stored representation.
The distinction matters for deployment: progressive rendering controls
computation at render time, whereas progressive compression controls
storage and transmission before rendering starts.
GoDe is complementary to these approaches and uses Octree-GS as one of
its evaluated backbones (Sec.~\ref{sec:experiments}).

\noindent\textbf{Compression of 3DGS.} Compression of neural scene representations gained prominence already
in the NeRF era, where hybrid explicit-implicit models~\cite{muller2022instant, sun2022direct, chen2022tensorf} traded memory efficiency for rendering speed, motivating early compression work targeting their explicit components~\cite{deng2023compressing, boostnerf}.
The shift to splat-based representations renewed these concerns at larger scale.
A large body of work addresses single-rate compression of splat-based representations, including pruning-based methods~\cite{hanson2025pup,fan2023lightgaussian},
entropy coding approaches~\cite{chen2024hac, chen2025hac++, wang2024contextgs, zhan2025cat, liu2024hemgs, ali2024elmgs, liu2024compgs, wang2024end, papantonakis2024reducing, ali2024trimming, yang2026lightweight, girish2024eaglesefficientaccelerated3d}. These methods form the current frontier of single-rate codecs and
serve as per-rate reference points in our evaluation. They all share the
same assumption: the optimized output corresponds to one point on the
rate--distortion curve, and supporting multiple budgets requires training
and storing multiple models independently.

\noindent\textbf{Progressive 3DGS compression.}
PCGS~\cite{chen2026pcgs} builds on HAC++~\cite{chen2025hac++} and Scaffold-GS to
achieve scalable compression via an anchor hierarchy, progressive
quantization, and a cross-level entropy context conditioned on anchor
feature codes.
CSGaussian~\cite{tseng2026csgaussian} pursues scalable progression with an
additional semantic segmentation objective and is therefore not directly
comparable to our radiance field compression setting.
Both methods achieve strong rate--distortion results within the
Scaffold-GS family.
Their strongest component (the entropy context conditioned on anchor
feature vectors) is also the least transferable: explicit Gaussians,
Beta kernels, Spherical Voronoi appearance, and triangle primitives do
not expose the same feature structure, so the context model has no
direct analog outside the anchor family.
GoDe takes the complementary approach: by relying only on gradient
magnitudes, simulated quantization, and dictionary-based entropy coding, it only requires differentiable rendering of primitive subsets, serializable
per-primitive parameters, and a per-backbone specification of parameter groups
and quantization rules.

\begin{table}[htbp]
\centering
\caption{\textbf{Method positioning.} Qualitative comparison of GoDe
against existing approaches.}
\label{tab:method_positioning}
\footnotesize
\renewcommand{\arraystretch}{1.05}
\resizebox{\columnwidth}{!}{
\begin{tabular}{lccccc}
\toprule
\textbf{Method} &
\textbf{Prog.\ render.} &
\textbf{Pruning} &
\textbf{Codec} &
\textbf{Prog.\ comp.} &
\textbf{Agnostic} \\
\midrule
\multicolumn{6}{c}{\textit{Progressive rendering / LoD -- no compressed bitstream}} \\
CLoD-GS         & \cmark & \xmark & \xmark & \xmark & \xmark \\
PRoGS           & \cmark & \xmark & \xmark & \xmark & \xmark \\
LapisGS         & \cmark & \xmark & \xmark & \xmark & \xmark \\
Octree-GS       & \cmark & \xmark & \xmark & \xmark & \xmark \\
Hierarchical-GS & \cmark & \xmark & \xmark & \xmark & \xmark \\
\midrule
\multicolumn{6}{c}{\textit{Single-rate compression -- explicit}} \\
PUP-3DGS        & \xmark & \cmark & \xmark & \xmark & \xmark \\
LightGaussian   & \xmark & \cmark & \cmark & \xmark & \xmark \\
RDO             & \xmark & \xmark & \cmark & \xmark & \xmark \\
Reduced-3DGS    & \xmark & \cmark & \xmark & \xmark & \xmark \\
Comp-GS         & \xmark & \cmark & \cmark & \xmark & \xmark \\
Codec-GS        & \xmark & \xmark & \cmark & \xmark & \xmark \\
\multicolumn{6}{c}{\textit{Single-rate compression -- neural decoder}} \\
HAC / HAC++     & \xmark & \cmark & \cmark & \xmark & \xmark \\
Context-GS      & \xmark & \cmark & \cmark & \xmark & \xmark \\
CAT-3DGS        & \xmark & \cmark & \cmark & \xmark & \xmark \\
HEMGS           & \xmark & \xmark & \cmark & \xmark & \xmark \\
\midrule
\multicolumn{6}{c}{\textit{Progressive compression -- Scaffold-GS family}} \\
PCGS            & \cmark & \cmark & \cmark & \cmark & \xmark \\
CSGaussian      & \cmark & \cmark & \cmark & \cmark & \xmark \\
\midrule
\textbf{GoDe}   & \cmark & \cmark & \cmark & \cmark & \cmark \\
\bottomrule
\end{tabular}
}
\end{table}

Tab.~\ref{tab:method_positioning} summarizes this landscape.
Existing works provide either progressive \emph{rendering} without a
scalable compressed representation, single-rate \emph{compression}
requiring one model per bitrate, or progressive \emph{compression}
within the Scaffold-GS anchor family.
GoDe is the only method in the table that combines a progressive
compressed bitstream with demonstrated support for substantially
different splat parameterizations.

\section{Method}
\label{sec:method}

\begin{figure*}[tp]
    \centering
    \includegraphics[width=\linewidth]{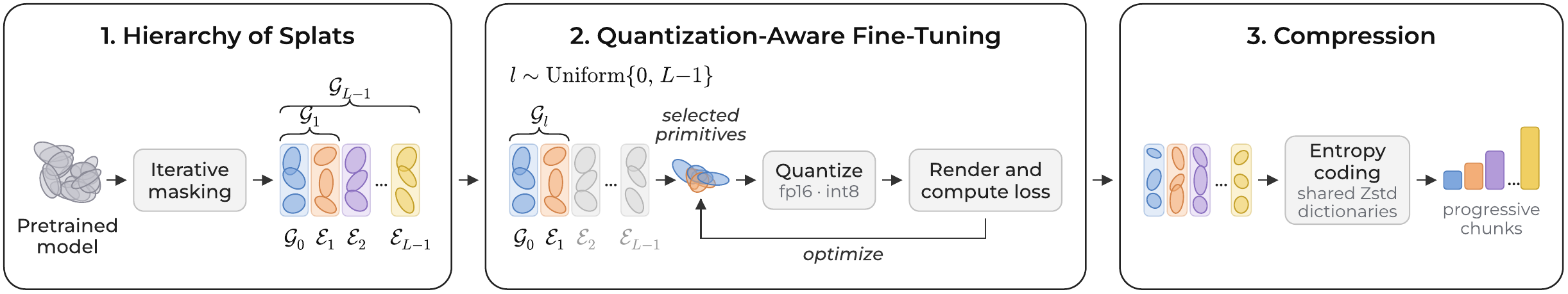}
    \caption{\textbf{GoDe pipeline.} From a trained representation:
    (1)~primitives are organized into a nested LoD hierarchy via
    iterative gradient-informed ranking; (2)~all LoDs are jointly
    fine-tuned under simulated quantization, sampling one LoD per
    iteration; (3)~the base layer and each enhancement layer are
    entropy-coded as independent chunks. }
    \label{fig:pipeline}
\end{figure*}

GoDe takes a trained splat-based representation and produces a single
progressive bitstream from which any prefix can be decoded and rendered
directly, without retraining or re-optimization.
The pipeline has three stages (Fig.~\ref{fig:pipeline}).
First, primitives are ranked by gradient-informed importance and
organized into a nested level-of-detail hierarchy, so that each coarser
level is a strict subset of every finer one
(Sec.~\ref{sec:hierarchy}).
Second, all levels are jointly fine-tuned under simulated quantization
in a single training stage, adapting the shared parameters to every
decoded prefix at once (Sec.~\ref{sec:finetuning}).
Third, the base layer and each enhancement layer are entropy coded as
independent chunks; decoding a level requires only concatenating the
chunks up to that level (Sec.~\ref{sec:coding}).
The pipeline treats primitives as atomic units and requires only a
differentiable renderer that supports rendering subsets of primitives,
a serializable per-primitive parameterization, and a per-backbone
specification of parameter groups and their quantization rules.
These requirements are met by all splat backbones we consider, which
allows the same procedure to transfer across them.

\subsection{Progressive Hierarchy Construction}
\label{sec:hierarchy}

Let $\mathcal{S}=\{p_i\}_{i=1}^{N}$ denote the set of $N$ primitives
of the trained model, each with parameters $\theta_i$.
GoDe constructs a sequence of $L$ nested subsets
$\mathcal{G}_0\subset\cdots\subset\mathcal{G}_{L-1}\subseteq\mathcal{S}$,
where $\mathcal{G}_0$ is the base layer and each finer level
$\mathcal{G}_l$ extends the previous one by a disjoint enhancement
layer $\mathcal{E}_l = \mathcal{G}_l \setminus \mathcal{G}_{l-1}$:
\begin{equation}
\mathcal{G}_l = \mathcal{G}_0 \cup \bigcup_{j=1}^{l} \mathcal{E}_j,
\quad l=1,\dots,L-1.
\label{eq:nesting}
\end{equation}
This decomposition guarantees that decoding level $l$ requires only
the first $l{+}1$ chunks, each carrying primitives not present in any
coarser level, with no primitive transmitted twice.
Level sizes follow a geometric progression,
\begin{equation}
n_l=\left\lfloor
  n_{\min}\left(n_{\max}/n_{\min}\right)^{l/(L-1)}
\right\rfloor,\quad l=0,\dots,L-1,
\label{eq:progression}
\end{equation}
with $n_{\min}=|\mathcal{G}_0|$ and $n_{\max}=|\mathcal{G}_{L-1}|\le N$.
A geometric schedule yields smaller absolute increments at low rates,
where additional primitives typically provide larger marginal quality
gains, and larger increments at high rates, where the rate--distortion
curve typically flattens. The hierarchy is built top-down, starting from the full model.
At each step, parameters are frozen and the gradient of the
photometric loss with respect to each primitive's parameters is
accumulated over all training views. Let
$g_i^{(m)}=\|\partial\mathcal{L}/\partial\theta_i^{(m)}\|_2$ denote
the gradient norm for parameter group $m$ of primitive $p_i$. Since
different groups have different dimensionalities and numerical
scales, directly summing $g_i^{(m)}$ can make one group dominate the
ranking. We therefore normalize each group by its root-mean-square
gradient over the current active set $\mathcal{A}$:
\begin{equation}
 r_m=\max\!\left(\epsilon,
 \sqrt{\frac{1}{|\mathcal{A}|}\sum_{p_j\in\mathcal{A}}
 \left(g_j^{(m)}\right)^2}\right),\qquad
 s_i=\sum_{m}\frac{g_i^{(m)}}{r_m},
\label{eq:importance}
\end{equation}
where $\epsilon>0$ is a small numerical constant.
The sum runs over all parameter groups of $p_i$. Including all groups is
important because a primitive can be critical through geometry, opacity,
kernel shape, or appearance; restricting the score to a single group
risks discarding primitives that are negligible in one attribute but
essential in another. The normalization places the group-wise scores on
comparable empirical scales before aggregation.
We first evaluate Eq.~\eqref{eq:importance} on the full model and set
$\mathcal{G}_{L-1}=\{p_i\in\mathcal{S}:s_i\geq\tau\}$ with
$\tau=10^{-12}$, used as a numerical cutoff for near-zero importance.
Thus, $n_{\max}=|\mathcal{G}_{L-1}|$ is scene-adaptive by removing only
primitives with numerically zero importance. Starting from
$\mathcal{G}_{L-1}$, we then iteratively remove the least important
primitives to obtain each coarser level.
At level $l$, we remove the $n_l-n_{l-1}$ primitives with the lowest
$s_i$ and assign them to the enhancement layer $\mathcal{E}_l$, yielding
$\mathcal{G}_{l-1} = \mathcal{G}_l \setminus \mathcal{E}_l$.
After each masking step, gradients are recomputed on the updated
active set before proceeding to the next coarser level.
Each step requires one gradient-accumulation sweep with no optimizer
update, so building the full $L$-level hierarchy takes approximately
$1$--$2$ minutes.

\subsection{Quantization-Aware Multi-Level Fine-Tuning}
\label{sec:finetuning}

Once the hierarchy is fixed, GoDe maintains a single global set of
primitive parameters $\Theta=\{\theta_i:p_i\in\mathcal{G}_{L-1}\}$.
Each level $\mathcal{G}_l$ renders the corresponding subset of
$\Theta$; whenever a primitive appears in multiple levels, its
attributes are shared rather than stored or optimized separately for
each LoD. Fine-tuning must therefore adapt these shared parameters to
perform well across all decoded prefixes and under the quantization
applied at coding time. A straightforward approach would accumulate the loss over all $L$
levels at each iteration, but this requires $L$ forward--backward
passes per step and makes training cost grow linearly with $L$.
Instead, GoDe samples one level
$l\sim\mathrm{Uniform}\{0,\dots,L-1\}$ and one training view $(c,I_c)$
per iteration, where $c$ denotes the camera and $I_c$ its reference
image. It renders only $\mathcal{G}_l$ with quantized parameters and takes one stochastic gradient step toward the objective
\begin{equation}
\min_{\Theta}\;\mathbb{E}_{l,c}\big[\mathcal{L}\big(\mathcal{R}(c,Q(\Theta_l)),I_c\big)
+\lambda\|\Theta_l^{(a)}\|_1\big],
\label{eq:qat}
\end{equation}
where $\Theta_l=\{\theta_i:p_i\in\mathcal{G}_l\}$ denotes the parameters of
level~$l$, $\|\Theta_l^{(a)}\|_1=\sum_{p_i\in\mathcal{G}_l}\|\theta_i^{(a)}\|_1$,
$Q(\cdot)$ simulates deployment quantization via a straight-through estimator
($\partial Q/\partial\theta\approx I$), and $\lambda$ controls the appearance
sparsity penalty. Uniform sampling makes Eq.~\eqref{eq:qat} the equally weighted mean of
the $L$ level-specific objectives. Thus, all levels are optimized jointly, while each iteration uses only one
LoD forward--backward pass, regardless of the number of levels $L$.
Full pseudocode is provided in Supp.\ Sec.~\ref{supp:pseudocode}.

\noindent\textbf{Quantization.}
We use two storage classes: geometric attributes are quantized to
\texttt{float16}, while appearance attributes are quantized to \texttt{int8}. The exact
attribute partition is detailed in
Supp.\ Sec.~\ref{supp:qat}.

\noindent\textbf{Sparsity regularization.}
Appearance attributes are typically among the most expensive
components to store in 3DGS-style representations.
We therefore apply the $\ell_1$ penalty in
Eq.~\eqref{eq:qat} to encourage sparsity in the appearance attributes
of the active primitives.

\subsection{Progressive Coding}
\label{sec:coding}

The disjoint structure of Eq.~\eqref{eq:nesting} means that
$\mathcal{G}_0$ and each $\mathcal{E}_l$ can be quantized and
serialized independently.
We compress each chunk with Zstandard using two shared dictionaries,
one for geometry and one for appearance, trained once on the
fine-tuned model and reused across all chunks.
Decoding level $l$ is then simply
$\mathrm{Dec}(l)=\bigcup_{j=0}^{l}\mathrm{Dec}(z_j)$, where $z_j$ denotes
the compressed chunk for $\mathcal{G}_0$ ($j{=}0$) or $\mathcal{E}_j$
($j{\ge}1$).
Receiving an enhancement layer at runtime triggers only decompression
and concatenation.

\subsection{Adaptive Deployment and Smooth Transitions}
\label{sec:deployment}

At deployment time, the active level is selected according to the
available budget: for a memory constraint $B$, one picks
$l^\star=\max\{l:\mathrm{Size}(z_0,\dots,z_l)\le B\}$; the same
prefix selection mechanism can use a measured rendering cost budget,
such as a target FPS. Since each decoded prefix is a complete,
renderable model, switching levels requires only decoding the next
chunk and appending its primitives to the active representation.
For interactive applications, switching levels abruptly activates a
full enhancement layer at once, which can produce a visible pop.
To mitigate this, we ramp the opacities of newly introduced
primitives when activating $\mathcal{E}_{l+1}$:
\begin{equation}
o_i(\alpha)=\alpha\,o_i,\quad\alpha\in[0,1],\;p_i\in\mathcal{E}_{l+1},
\label{eq:opacity-interp}
\end{equation}
so that rendering $\mathcal{G}_l\cup\alpha\mathcal{E}_{l+1}$ produces
a smooth transition without altering the stored bitstream.
\section{Experiments}
\label{sec:experiments}

\noindent\textbf{Setup.}
We evaluate on Mip-NeRF360~\cite{barron2022mip},
Tanks\&Temples~\cite{knapitsch2017tanks}, and DeepBlending~\cite{DeepBlending2018}, reporting PSNR, SSIM, LPIPS~(VGG),
size~(MB), and average test-set FPS on a single NVIDIA A40.
The same GoDe pipeline is applied to five backbone configurations
spanning the primitive, kernel, and appearance choices discussed in
Sec.~\ref{sec:related}: Scaffold-GS~\cite{scaffoldgs}, 3DGS with MCMC-based
training~\cite{kheradmand2024d} and spherical-harmonics appearance, BS+SV
(Beta Splatting~\cite{liu2025deformable} with spherical Voronoi~\cite{di2026spherical}
appearance), Octree-GS~\cite{ren2024octree}, and Triangle
Splatting~\cite{held2026triangle}.
For every model, we use $L{=}8$ levels and uniform level sampling.
Hierarchy endpoints, fine-tuning schedules, and per-backbone settings
are detailed in supp.\ Sec.~\ref{supp:qat}.

\subsection{Main Comparison}
\label{sec:main_comparison}

\begin{table*}[tp]
\centering
\caption{
\textbf{Main quantitative comparison.}
Each row of a non-scalable method corresponds to an independently
optimized operating point, whereas scalable methods obtain multiple
rates from a single trained model. For scalable methods, we show three
representative decoded prefixes spanning the range of competing
operating points; rate labels denote the selected operating points,
not LoD indices. GoDe uses $8$ levels in the main experiments.
$\dagger$ denotes methods whose public implementations were unavailable
at submission time.
}
\label{tab:main_results}
\scriptsize
\setlength{\tabcolsep}{2.35pt}
\renewcommand{\arraystretch}{0.98}
\resizebox{0.9\textwidth}{!}{%
\begin{tabular}{l c | *{5}{c} | *{5}{c} | *{5}{c}}
\toprule
\multirow{2}{*}{Method} & \multirow{2}{*}{Rate} &
\multicolumn{5}{c}{Mip-NeRF360} &
\multicolumn{5}{c}{Tanks\&Temples} &
\multicolumn{5}{c}{DeepBlending} \\
\cmidrule(lr){3-7}
\cmidrule(lr){8-12}
\cmidrule(l){13-17}
& &
PSNR$\uparrow$ & SSIM$\uparrow$ & LPIPS$\downarrow$ & Size [MB]$\downarrow$ & FPS$\uparrow$ &
PSNR$\uparrow$ & SSIM$\uparrow$ & LPIPS$\downarrow$ & Size [MB]$\downarrow$ & FPS$\uparrow$ &
PSNR$\uparrow$ & SSIM$\uparrow$ & LPIPS$\downarrow$ & Size [MB]$\downarrow$ & FPS$\uparrow$ \\
\midrule

\rowcolor{black!5}
\multicolumn{17}{c}{\textit{Non-scalable -- Neural}} \\

\multirow{2}{*}{HEMGS$^\dagger$~\cite{liu2024hemgs}}
 & Low
 & 27.75 & 0.806 & 0.248 & 12.50 & --
 & 24.42 & 0.848 & 0.192 & 6.00  & --
 & 30.24 & 0.908 & 0.266 & 3.00  & -- \\
 & High
 & 27.93 & 0.813 & 0.230 & 21.00 & --
 & 24.58 & 0.856 & 0.176 & 10.10 & --
 & 30.37 & 0.911 & 0.253 & 6.70  & -- \\

\multirow{2}{*}{HAC++~\cite{chen2025hac++}}
 & Low
 & 27.60 & 0.803 & 0.253 & 8.34  & 70
 & 24.22 & 0.849 & 0.190 & 5.18  & 79
 & 30.16 & 0.907 & 0.266 & 2.91  & 152 \\
 & High
 & 27.82 & 0.811 & 0.231 & 18.48 & 71
 & 24.32 & 0.854 & 0.178 & 8.63  & 78
 & 30.34 & 0.911 & 0.254 & 5.28  & 154 \\

\multirow{2}{*}{Context-GS~\cite{wang2024contextgs}}
 & Low
 & 27.62 & 0.808 & 0.237 & 12.68 & 66
 & 24.20 & 0.852 & 0.184 & 7.05  & 77
 & 30.11 & 0.907 & 0.265 & 3.45  & 135 \\
 & High
 & 27.75 & 0.811 & 0.231 & 18.41 & 58
 & 24.29 & 0.855 & 0.176 & 11.80 & 78
 & 30.39 & 0.909 & 0.258 & 6.60  & 121 \\

\multirow{2}{*}{Cat-3DGS~\cite{zhan2025cat}}
 & Low
 & 25.82 & 0.730 & 0.362 & 1.72  & 73
 & 22.97 & 0.786 & 0.293 & 1.42  & 82
 & 28.53 & 0.878 & 0.336 & 0.93  & 161 \\
 & High
 & 27.77 & 0.809 & 0.241 & 12.35 & 74
 & 24.41 & 0.853 & 0.189 & 6.93  & 83
 & 30.29 & 0.909 & 0.269 & 3.56  & 161 \\

\methodrule

\rowcolor{black!5}
\multicolumn{17}{c}{\textit{Non-scalable -- Explicit}} \\

Codec-GS$^\dagger$~\cite{yang2026lightweight}
 & --
 & 27.30 & 0.810 & 0.236 & 10.30 & --
 & 23.63 & 0.841 & 0.192 & 7.80  & --
 & 29.81 & 0.906 & 0.251 & 9.00  & -- \\

\multirow{2}{*}{RDO~\cite{wang2024end}}
 & Low
 & 26.45 & 0.782 & 0.322 & 9.70  & 280
 & 23.19 & 0.827 & 0.253 & 5.50  & 370
 & 29.61 & 0.903 & 0.331 & 7.00  & 335 \\
 & High
 & 27.05 & 0.801 & 0.288 & 23.30 & 178
 & 23.32 & 0.839 & 0.232 & 11.90 & 257
 & 29.72 & 0.906 & 0.318 & 18.00 & 200 \\

\multirow{2}{*}{Reduced-3DGS~\cite{papantonakis2024reducing}}
 & Low
 & 27.05 & 0.807 & 0.272 & 22.70 & 247
 & 23.46 & 0.840 & 0.228 & 10.50 & 384
 & 29.63 & 0.906 & 0.318 & 13.60 & 337 \\
 & High
 & 27.28 & 0.813 & 0.264 & 45.80 & 240
 & 23.57 & 0.844 & 0.221 & 20.70 & 360
 & 29.69 & 0.907 & 0.314 & 35.30 & 311 \\

\multirow{2}{*}{Comp-GS~\cite{liu2024compgs}}
 & Low
 & 26.40 & 0.778 & 0.323 & 9.10  & 233
 & 23.06 & 0.816 & 0.278 & 6.10  & 334
 & 29.22 & 0.894 & 0.369 & 6.00  & 440 \\
 & High
 & 27.28 & 0.802 & 0.283 & 36.40 & 165
 & 23.62 & 0.837 & 0.248 & 23.10 & 286
 & 29.89 & 0.904 & 0.336 & 23.10 & 254 \\

LightGaussian~\cite{fan2023lightgaussian}
 & --
 & 27.24 & 0.810 & 0.273 & 51.00 & 161
 & 23.55 & 0.839 & 0.235 & 28.50 & 225
 & 29.41 & 0.904 & 0.329 & 43.20 & 348 \\

\methodrule

\rowcolor{black!5}
\multicolumn{17}{c}{\textit{Scalable -- Neural}} \\

\multirow{3}{*}{PCGS~\cite{chen2026pcgs}}
 & R1
 & 27.82 & 0.809 & 0.292 & 12.20 & 66
 & 24.32 & 0.849 & 0.235 & 4.93  & 107
 & 30.15 & 0.906 & 0.344 & 2.84  & 147 \\
 & R2
 & 27.93 & 0.811 & 0.288 & 16.26 & 67
 & 24.43 & 0.852 & 0.230 & 6.62  & 105
 & 30.27 & 0.908 & 0.340 & 3.81  & 146 \\
 & R3
 & 27.96 & 0.812 & 0.287 & 20.84 & 66
 & 24.45 & 0.853 & 0.228 & 8.79  & 104
 & 30.30 & 0.909 & 0.339 & 5.12  & 143 \\

\specialrule{0.18pt}{0.25pt}{0.25pt}

\multirow{3}{*}{\makecell[l]{\textbf{GoDe}\\\textbf{(Scaffold-GS)}}}
 & R1
 & 27.62 & 0.802 & 0.297 & 11.93 & 96
 & 23.68 & 0.829 & 0.259 & 5.10  & 160
 & 30.00 & 0.901 & 0.351 & 3.92  & 183 \\
 & R2
 & 27.87 & 0.811 & 0.282 & 17.85 & 83
 & 24.16 & 0.847 & 0.229 & 7.80  & 137
 & 30.27 & 0.905 & 0.343 & 4.63  & 178 \\
 & R3
 & 27.92 & 0.813 & 0.278 & 21.95 & 78
 & 24.25 & 0.851 & 0.220 & 9.66  & 125
 & 30.40 & 0.908 & 0.337 & 5.48  & 171 \\

\methodrule

\rowcolor{black!5}
\multicolumn{17}{c}{\textit{Scalable -- Explicit}} \\

\multirow{3}{*}{\makecell[l]{\textbf{GoDe}\\\textbf{(3DGS)}}}
 & R1
 & 26.97 & 0.781 & 0.322 & 11.31 & 270
 & 23.12 & 0.804 & 0.297 & 5.44  & 421
 & 29.01 & 0.885 & 0.378 & 4.25  & 562 \\
 & R2
 & 27.50 & 0.807 & 0.286 & 18.16 & 220
 & 23.51 & 0.819 & 0.274 & 7.23  & 374
 & 29.41 & 0.893 & 0.362 & 5.94  & 486 \\
 & R3
 & 27.60 & 0.813 & 0.274 & 23.21 & 198
 & 23.79 & 0.829 & 0.254 & 9.71  & 324
 & 29.58 & 0.897 & 0.350 & 8.37  & 429 \\

\specialrule{0.12pt}{0.18pt}{0.18pt}

\multirow{3}{*}{\makecell[l]{\textbf{GoDe}\\\textbf{(BS+SV)}}}
 & R1
 & 27.53 & 0.785 & 0.323 & 9.97 & 213
 & 23.62 & 0.814 & 0.285 & 4.74  & 357
 & 29.70 & 0.892 & 0.370 & 4.80  & 365 \\
 & R2
 & 28.11 & 0.811 & 0.283 & 16.32 & 171
 & 24.08 & 0.830 & 0.260 & 6.15  & 328
 & 29.96 & 0.899 & 0.355 & 6.61  & 323 \\
 & R3
 & 28.23 & 0.818 & 0.269 & 20.96 & 152
 & 24.37 & 0.841 & 0.240 & 8.00  & 301
 & 30.08 & 0.902 & 0.344 & 9.10  & 289 \\

\bottomrule
\end{tabular}%
}
\end{table*}

\begin{figure*}[t]
\centering
\includegraphics[width=0.85\linewidth]{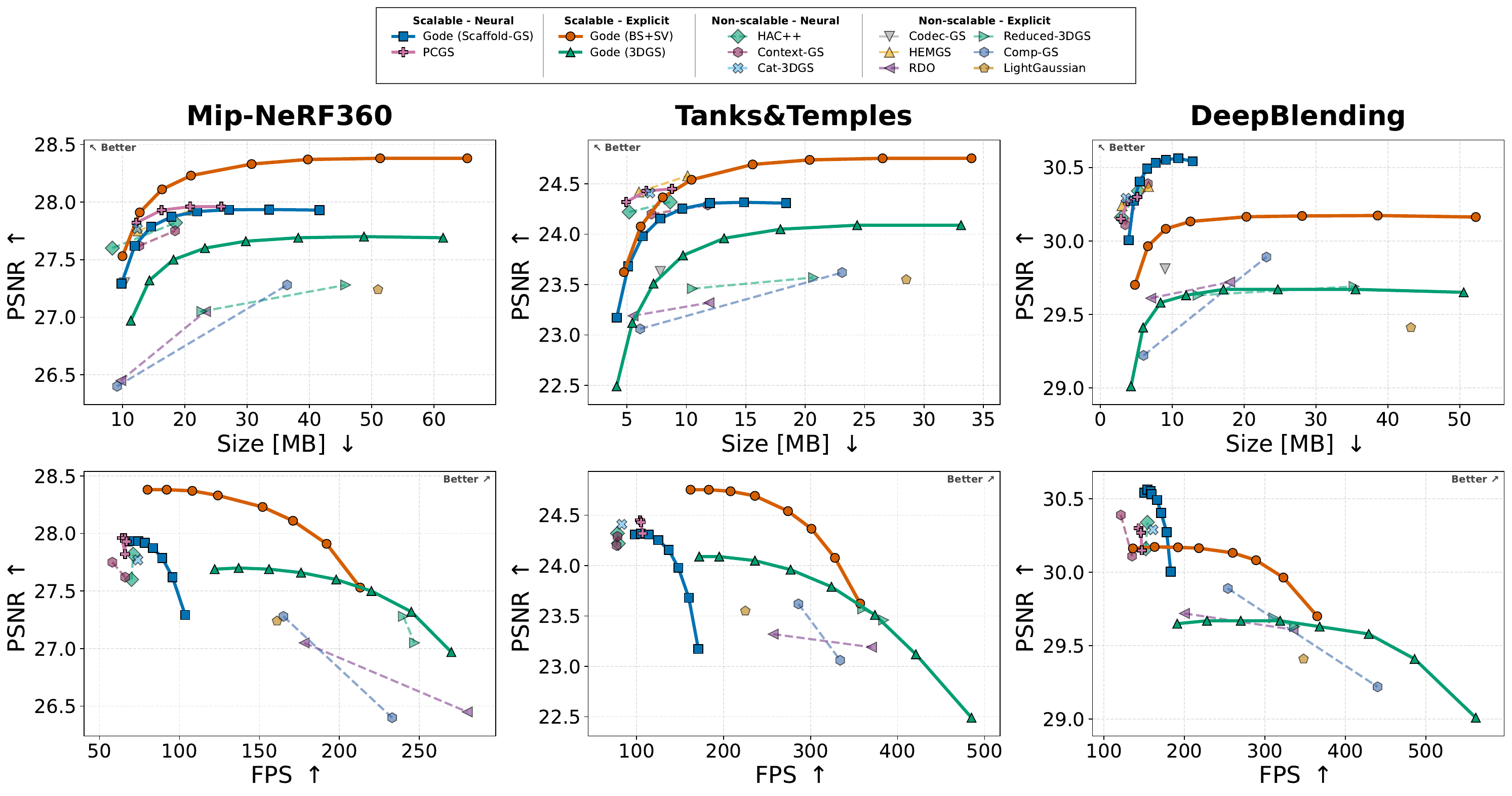}
\caption{\textbf{Rate--distortion and speed--distortion curves.}
Each point is averaged over the dataset; solid lines denote scalable
methods, dashed lines denote non-scalable ones.}
\label{fig:main_plot}
\end{figure*}

Table~\ref{tab:main_results} and Fig.~\ref{fig:main_plot} summarize
rate--distortion, storage, and rendering speed. Each GoDe operating point
is obtained by decoding a prefix of a single progressive model, whereas
non-scalable baselines optimize a separate model for each target rate.

\noindent\textbf{Explicit backbones.}
Among explicit methods, GoDe is the only scalable approach in the
comparison. All competing baselines optimize each reported operating
point independently for a specific rate, while GoDe exposes the entire
rate--distortion curve from a single trained model. Despite this stricter
constraint, GoDe (3DGS) remains competitive: on Mip-NeRF360, it reaches
$27.50$\,dB at $18.16$\,MB and $27.60$\,dB at $23.21$\,MB,
outperforming RDO, Reduced-3DGS, Comp-GS, and LightGaussian while
using comparable or substantially lower storage. Codec-GS achieves stronger isolated operating points
in parts of the range, but its public implementation is unavailable
for FPS evaluation under our protocol. The same progressive pipeline directly benefits from stronger explicit
representations. GoDe (BS+SV) reaches $28.11$\,dB at $16.32$\,MB on
Mip-NeRF360, further improving the rate--distortion curve without
changing the scalability mechanism. 

\noindent\textbf{Neural backbone.}
PCGS is the closest scalable baseline. GoDe (Scaffold-GS) remains competitive
on Mip-NeRF360 and DeepBlending, with PSNR differences within
$0.20$\,dB across the reported operating points and comparable SSIM on
Mip-NeRF360, while PCGS performs better on Tanks\&Temples. GoDe also
achieves lower LPIPS at the medium and high reported rates on both
Mip-NeRF360 and Tanks\&Temples. PCGS is specialized for the
Scaffold-GS/HAC++ representation and its learned entropy model, and
optimizes only $3$--$4$ discrete quality levels. In contrast, GoDe
applies the same progressive pipeline across heterogeneous backbones
and supports substantially finer rate granularity, using $8$ levels in
our main experiments and up to $32$ in our ablations. GoDe is also consistently faster, rendering on average $28\%$ faster
than PCGS across the reported datasets and operating points.

\subsection{Efficiency}
\label{sec:efficiency}

\begin{table}[htbp]
\caption{
\textbf{Training, encoding, and decoding time.} Training is in hours; coding is in seconds. $^\dagger$ PCGS uses one shared scalable encoding for all bitrate levels.
}
\label{tab:perf}
\centering
\footnotesize
\setlength{\tabcolsep}{4pt}
\renewcommand{\arraystretch}{1.08}
\resizebox{\columnwidth}{!}{%
\begin{tabular}{l c cc ccc ccc}
\toprule
Method
& Scalable
& \multicolumn{2}{c}{Train$\downarrow$}
& \multicolumn{3}{c}{Encoding$\downarrow$}
& \multicolumn{3}{c}{Decoding$\downarrow$} \\
\cmidrule(lr){3-4}
\cmidrule(lr){5-7}
\cmidrule(l){8-10}
& 
& Per run & All
& Low & Medium & High
& Low & Medium & High \\
\midrule
Context-GS & \xmark & 1.3 & 4.0 & 52.7 & 87.8 & 97.2 & 53.1 & 82.5 & 88.1 \\
HAC++      & \xmark & 1.3 & 3.9 & 8.9  & 10.1 & 16.4 & 12.3 & 13.4 & 19.0 \\
RDO     & \xmark & 1.3 & 3.9 & 1.1  & 2.9  & 4.8  & 2.2  & 7.4  & 19.0 \\
Comp-GS    & \xmark & 1.6 & 4.9 & 10.8 & 10.8 & 10.9 & 9.1  & 9.0  & 8.8  \\
\addlinespace[0.15cm]
PCGS$^\dagger$             & \cmark & 1.83 & 1.83 & \multicolumn{3}{c}{74.49} & 42.3 & 46.3 & 52.1 \\
GoDe (3DGS)                & \cmark & 1.1 & 1.1 & 2.1 & 2.9 & 4.2 & 0.6 & 0.9 & 1.3 \\
GoDe (Scaffold-GS)         & \cmark & 0.6 & 0.6 & 0.1 & 0.2 & 0.3 & 0.2 & 0.3 & 0.5 \\
\bottomrule
\end{tabular}%
}
\vspace{0.05cm}
\scriptsize
\end{table}

Table~\ref{tab:perf} reports training, encoding, and cumulative decoding
time separately. GoDe decouples hierarchy construction and multi-LoD
optimization from the entropy coding backend. We therefore use
Zstandard as a lightweight backend whose low latency is well suited to
progressive streaming. The complete Scaffold-GS and 3DGS pipelines require $0.6$ and $1.1$\,h of training, respectively, compared with $1.8$\,h for PCGS. For the three reported operating points,
non-scalable baselines require $3.9$--$4.9$\,h of aggregate training,
since each rate is optimized separately. Encoding is substantially
faster as well: GoDe requires $0.1$--$0.3$\,s for Scaffold-GS and
$2.1$--$4.2$\,s for 3DGS, whereas PCGS encoding takes
$74.49$\,s. The largest deployment difference is cumulative decoding. GoDe requires only $0.2$--$0.5$\,s for Scaffold-GS and $0.6$--$1.3$\,s for 3DGS,
compared with $42.3$--$52.1$\,s for PCGS. This tens-of-seconds latency
limits the practicality of PCGS for responsive progressive streaming,
where enhancement layers may need to be activated on demand.

\subsection{Backbone Agnosticism}
\label{sec:agnostic}

\begin{figure*}
    \centering
    \includegraphics[width=\linewidth]{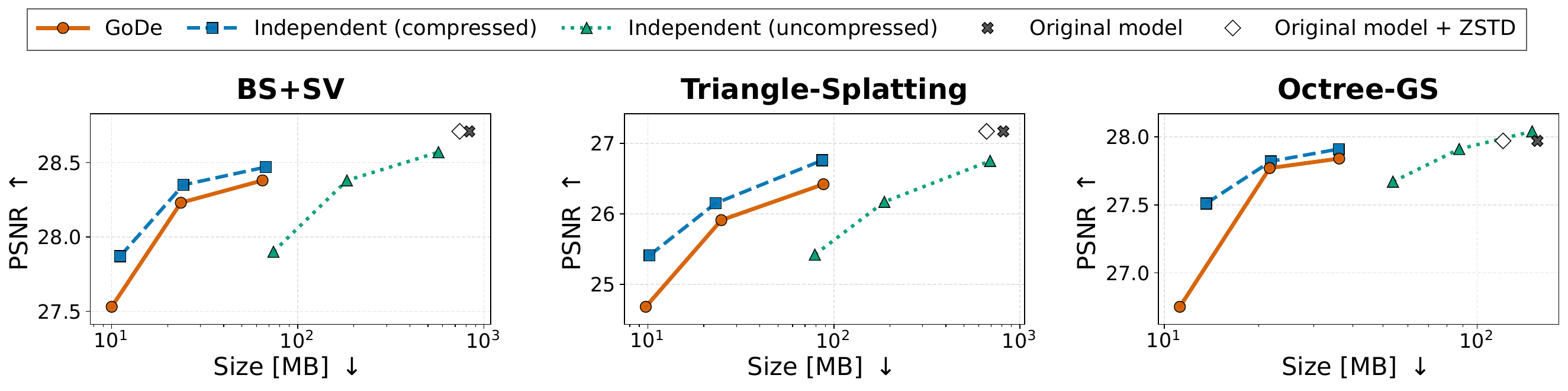}
    \caption{\textbf{Cross-backbone transfer on Mip-NeRF360.}
   GoDe and independently optimized same-backbone models matched
    by primitive count. Results are averaged over the nine scenes.}
    \label{fig:transferability}
\end{figure*}

For BS+SV, Triangle Splatting, and Octree-GS, no established
same-backbone compression baselines are available. We therefore train
independent models at the primitive budgets of GoDe LoDs 2, 4, and 7,
optimizing each operating point separately for $30$k iterations.
We report these controls before and after applying the same quantization
and compression pipeline used for GoDe. For reference, we also report
the original, unmodified model and the same model after lossless
Zstandard compression only. Fig.~\ref{fig:transferability} shows PSNR--size results on Mip-NeRF360. Although each independent model can
specialize to a single rate, GoDe remains close while providing all
operating points from one progressive model. The same behavior across
BS+SV, Triangle Splatting, and Octree-GS supports transfer across
substantially different splat representations.

\subsection{Qualitative Results}
\label{sec:qualitative}

\begin{figure*}[t]
    \centering
    \includegraphics[width=\linewidth]{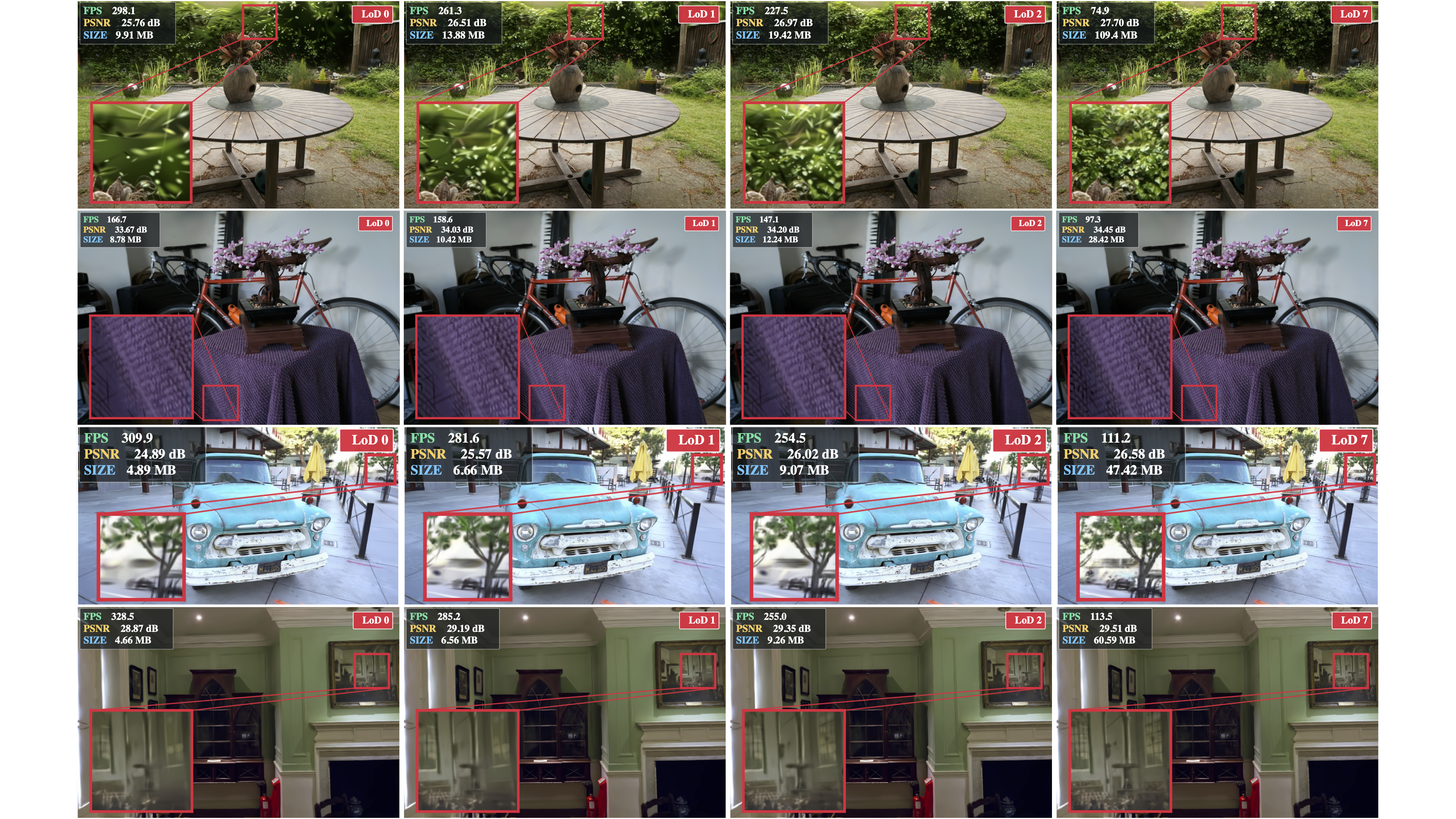}
    \caption{\textbf{Level-of-detail progression.} High-resolution
    crops show how coarse levels preserve global structure while
    enhancement layers progressively restore fine geometry and
    view-dependent appearance.}
    \label{fig:qualitative}
\end{figure*}

Fig.~\ref{fig:qualitative} shows successive decoded prefixes:
coarse levels preserve the scene structure, while enhancement layers
progressively restore fine geometry and appearance.
\begin{figure}[t]
    \centering
    \includegraphics[width=\linewidth]{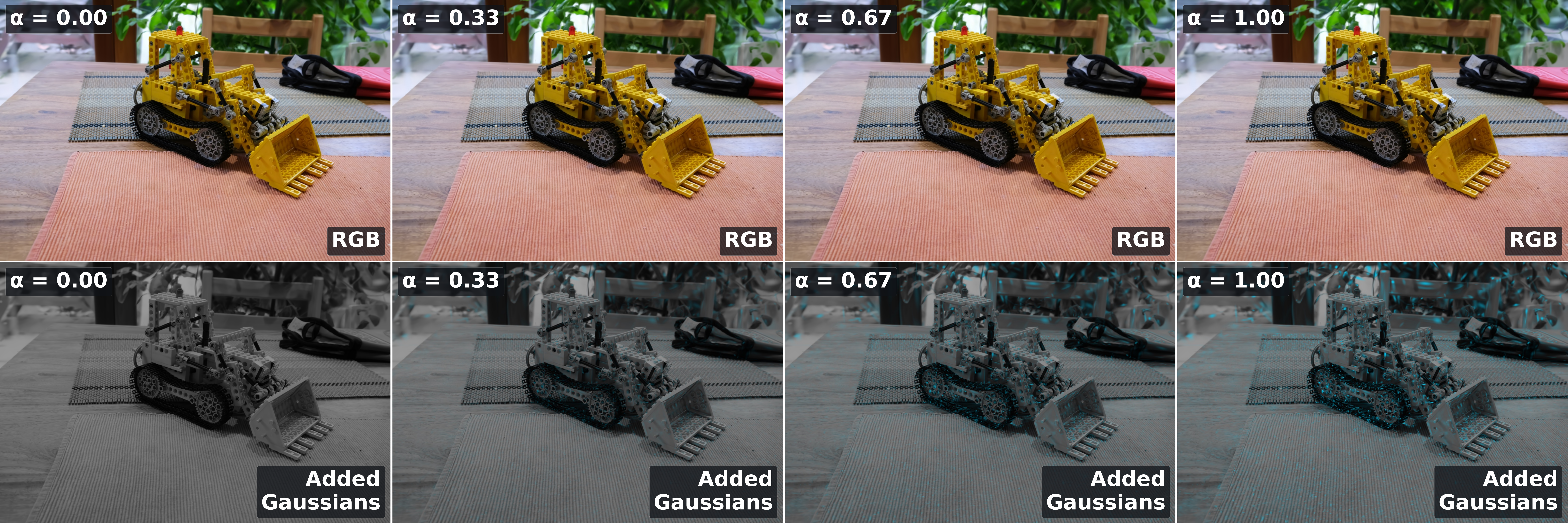}
    \caption{\textbf{Smooth level transitions.}
    Opacity interpolation progressively activates an enhancement layer,
    reducing visible popping between adjacent LoDs.}
    \label{fig:continuous}
\end{figure}
Fig.~\ref{fig:continuous} shows how the opacity ramp of
Eq.~\ref{eq:opacity-interp} progressively activates the primitives of
an enhancement layer and reduces popping between adjacent levels.

\subsection{Ablations}
\label{sec:ablations}

\begin{figure}[t]
    \centering
    \includegraphics[width=0.9\linewidth]{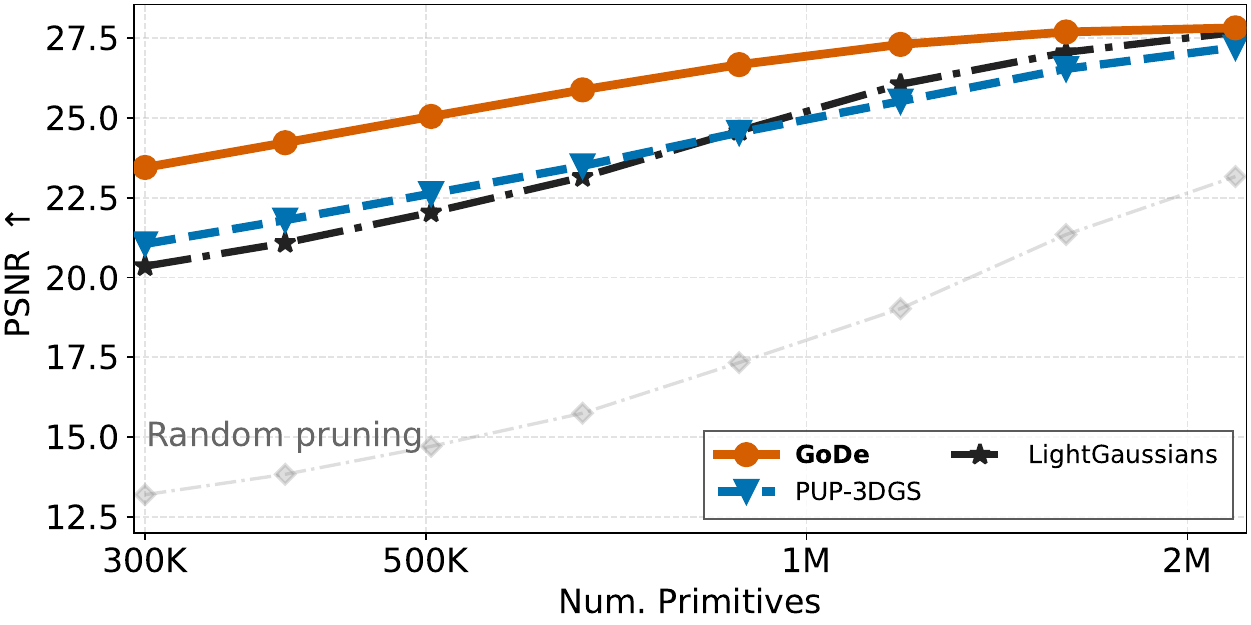}
    \caption{\textbf{Importance-criterion ablation on Mip-NeRF360.}
    Our criterion consistently preserves higher
    reconstruction quality at matched active primitive counts than
    established Gaussian-pruning criteria.}
    \label{fig:ablations}
\end{figure}

\noindent\textbf{Importance criterion.}
Fig.~\ref{fig:ablations} compares our iterative
masked-gradient score (Eq.~\ref{eq:importance}) with LightGaussian
and PUP-3DGS, two strong and widely used pruning criteria for explicit
Gaussian representations. All curves are averaged over the nine
Mip-NeRF360 scenes and evaluated at matched active primitive counts.
Each criterion is used within the same iterative hierarchy construction
and recomputed after every masking step, isolating the ranking score
from the remaining pipeline. Our normalized multi-parameter criterion
consistently achieves higher PSNR at matched primitive counts.
Supplementary Sec.~\ref{supp:score_normalization} provides the broader
diagnostic comparison with alternative score formulations.

\begin{figure}
    \centering
    \includegraphics[width=\linewidth]{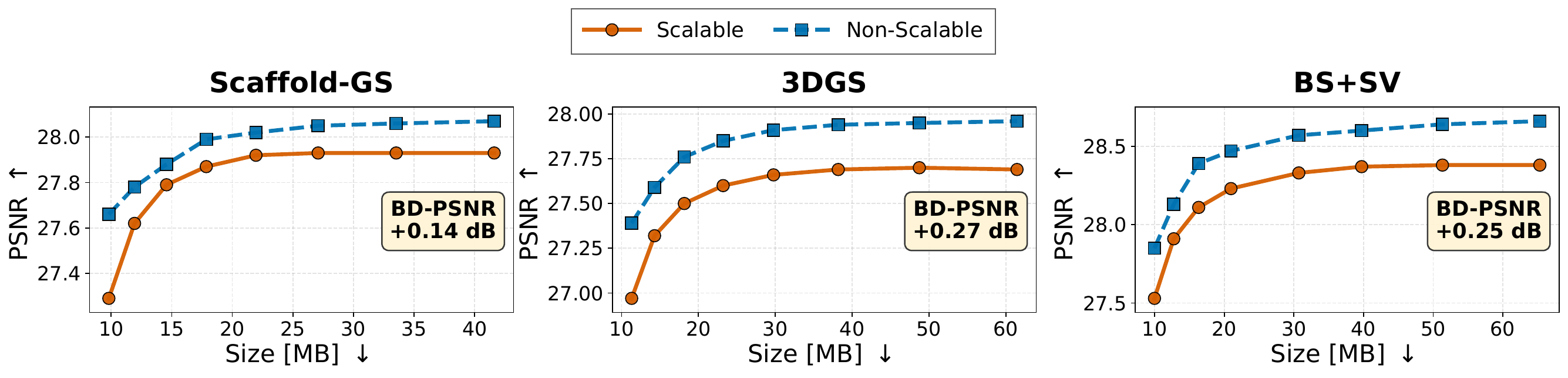}
    \caption{\textbf{Cost of scalability.} BD-PSNR on Mip-NeRF360
    comparing scalable GoDe with a matched-budget non-scalable variant.}
    \label{fig:hierarchical_vs_independent}
\end{figure}

\noindent\textbf{Cost of scalability.}
Fig.~\ref{fig:hierarchical_vs_independent} compares GoDe with a
non-scalable variant using the same pretrained model, ranking criterion,
target primitive counts, and total optimization budget. At each stage,
the non-scalable variant ranks, prunes to the next target, and fine-tunes
that model for $30$k/$L$ iterations before repeating the process.
With $L{=}8$, both methods therefore use the same total $30$k optimizer
steps. The non-scalable variant can specialize after each pruning stage,
yet improves BD-PSNR by only $0.14$\,dB for Scaffold-GS, $0.27$\,dB for
3DGS, and $0.25$\,dB for BS+SV. This gap quantifies the cost of sharing
one progressively decodable model across all rates.

\noindent\textbf{Impact of fine-tuning and compression.}
Table~\ref{tab:impact_ft_compression} isolates the contribution of each
pipeline stage on Mip-NeRF360 with 3DGS. Masking alone strongly degrades
coarse levels, and multi-level fine-tuning recovers $+3.61$\,dB at the
lowest rate. Quantization then reduces storage by $3.5\times$ at every
level for at most $0.11$\,dB, and lossless Zstandard coding adds a
further $1.8$--$2.4\times$.

\begin{table}[t]
\centering
\includegraphics[width=\columnwidth]{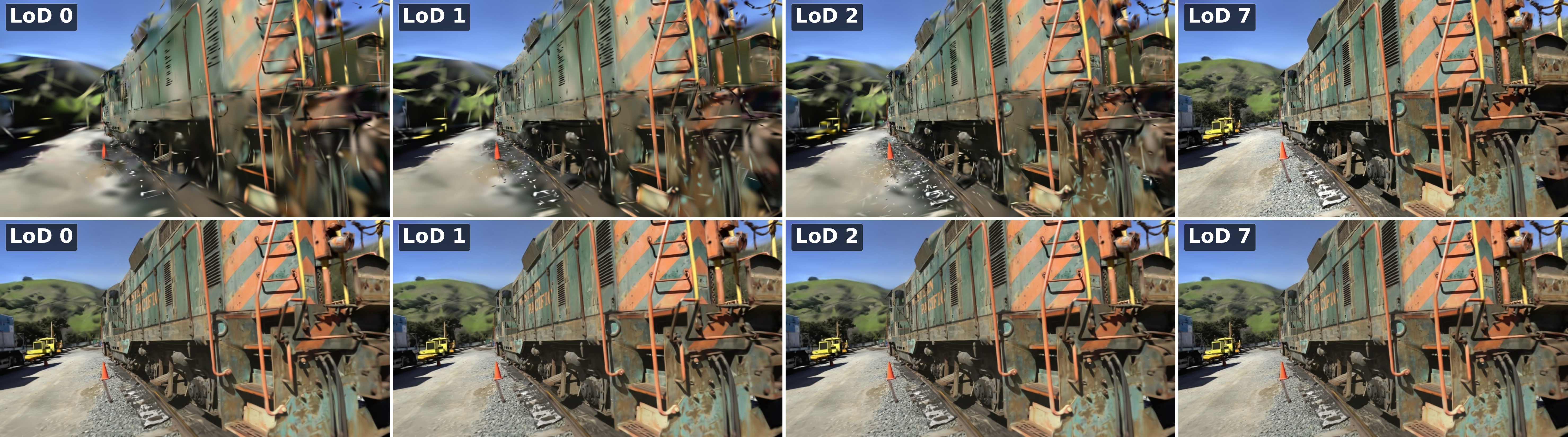}
\vspace{0.08cm}

\footnotesize
\setlength{\tabcolsep}{3.2pt}
\renewcommand{\arraystretch}{1.05}
\resizebox{\columnwidth}{!}{
\begin{tabular}{c c cc cc cc c}
\toprule
& &
\multicolumn{2}{c}{\textit{Mask only}}
& \multicolumn{2}{c}{\textit{$+$ Fine-tuning}}
& \multicolumn{2}{c}{\textit{$+$ Quantization}}
& \textit{$+$ ZSTD } \\
\cmidrule(lr){3-4}
\cmidrule(lr){5-6}
\cmidrule(lr){7-8}
\cmidrule(lr){9-9}
Rate & Retained
& PSNR $\uparrow$ & Size [MB] $\downarrow$
& PSNR $\uparrow$ & Size [MB] $\downarrow$
& PSNR $\uparrow$ & Size [MB] $\downarrow$
& Size [MB] $\downarrow$ \\
\midrule
0 & 0.13 & 23.45 & 70.95  & 27.06 & 70.95  & 26.97 & 20.03  & 11.31 \\
1 & 0.16 & 24.22 & 91.57  & 27.41 & 91.57  & 27.32 & 25.85  & 14.29 \\
2 & 0.19 & 25.05 & 119.42 & 27.60 & 119.42 & 27.50 & 33.71  & 18.16 \\
3 & 0.24 & 25.88 & 157.32 & 27.70 & 157.32 & 27.60 & 44.40  & 23.21 \\
4 & 0.30 & 26.67 & 209.19 & 27.77 & 209.19 & 27.66 & 59.05  & 29.77 \\
5 & 0.38 & 27.30 & 280.56 & 27.80 & 280.56 & 27.69 & 79.19  & 38.19 \\
6 & 0.49 & 27.69 & 379.23 & 27.81 & 379.23 & 27.70 & 107.04 & 48.74 \\
7 & 0.63 & 27.82 & 516.17 & 27.80 & 516.17 & 27.69 & 145.69 & 61.45 \\
\bottomrule
\end{tabular}
}
\caption{\textbf{Impact of fine-tuning, quantization and entropy coding} 
on Mip-NeRF360 with 3DGS.
Top: masked LoDs before and after
fine-tuning. Bottom: quantitative results and final compression.
\textit{Retained}: fraction of Gaussians of the full model.
}
\label{tab:impact_ft_compression}
\end{table}

\noindent\textbf{Additional analyses.} Supplementary Sec.~\ref{supp:additional_ablations} covers ranking, LoD and bound ablations, PUP-3DGS comparison, convergence, and seed robustness.
\section{Conclusion}
\label{sec:conclusion}
We presented GoDe, a progressive compression framework for splat-based
radiance fields: starting from a trained model, it ranks primitives by
gradient-informed importance, applies quantization-aware fine-tuning
across all levels, and entropy-codes the resulting layers for progressive decoding.
Across five heterogeneous backbones, the same pipeline produces monotonic rate--distortion curves and competitive quality while enabling
fast rendering on explicit representations; most importantly for
deployment, cumulative decoding takes  $0.2$--$1.3$\,s, one to two
orders of magnitude faster than existing methods.
These results indicate that progressive delivery can be added to
existing splat-based pipelines without sacrificing their interactive
rendering advantages, making GoDe suitable for bandwidth-adaptive
streaming and scalable storage of radiance fields.

\noindent\textbf{Limitations.}
The proposed hierarchy introduces a minor coding overhead, though this cost is small relative to the
benefit of progressive access; moreover, since GoDe is deliberately model-agnostic, architecture-specific designs could still improve individual backbones further. 
Finally, like other optimization-based compression methods, GoDe assumes the availability of a training set.



\newpage
{
    \small
    \bibliographystyle{ieeenat_fullname}
    \bibliography{main}
}

\clearpage
\setcounter{section}{0}
\renewcommand{\thesection}{S\arabic{section}}
\renewcommand{\thetable}{S\arabic{table}}
\renewcommand{\thefigure}{S\arabic{figure}}
\renewcommand{\theHsection}{supp.\arabic{section}}
\renewcommand{\theHtable}{supp.\arabic{table}}
\renewcommand{\theHfigure}{supp.\arabic{figure}}
\setcounter{table}{0}
\setcounter{figure}{0}

\begin{center}
{\Large\bf GoDe: Supplementary Material}
\end{center}
\vspace{0.5em}

\noindent
This supplementary material provides additional implementation details,
ablations, and quantitative and qualitative results for GoDe.
Sec.~\ref{supp:pseudocode} provides pseudocode for hierarchy
construction, multi-level QAT, and progressive coding.
Sec.~\ref{supp:additional_ablations} studies score normalization,
the number of levels, hierarchy bounds, the PUP-3DGS pruning protocol,
convergence, and seed robustness.
Sec.~\ref{supp:backbone_controls} extends the cross-backbone transfer
analysis to all three datasets using independently optimized reference
models.
Sec.~\ref{supp:qat} details the quantization scheme, fine-tuning
objective, and per-backbone settings.
Finally, Sec.~\ref{supp:fulltables} reports the complete per-level
quantitative results and the full eight-level LoD progression across
the three benchmarks.

\section{Pseudocode}
\label{supp:pseudocode}

GoDe operates on a trained representation whose primitives can be
activated or removed as groups.
The hierarchy is stored as a base set $\mathcal{G}_0$ and ordered
enhancement sets $\{\mathcal{E}_1,\dots,\mathcal{E}_{L-1}\}$.
Algorithm~\ref{alg:hierarchy} builds these sets top-down by iterative
masked-gradient ranking.
Algorithm~\ref{alg:qat} jointly fine-tunes all decoded prefixes under
simulated quantization using stochastic level sampling.
Algorithm~\ref{alg:coding} encodes the resulting hierarchy into
independent progressive chunks.

\begin{algorithm}[h]
\caption{Iterative hierarchy construction}
\label{alg:hierarchy}
\begin{algorithmic}
\Require Trained representation $\mathcal{S}$, training views
  $\mathcal{D}$, lower level size $n_{\min}$, threshold $\tau$
\Ensure Base layer $\mathcal{G}_0$ and enhancement layers
  $\{\mathcal{E}_l\}_{l=1}^{L-1}$
\State Freeze $\Theta$ and accumulate gradients of $\mathcal{L}$
  over all views in $\mathcal{D}$ on $\mathcal{S}$
\ForAll{parameter groups $m$}
    \State $g_i^{(m)} \gets
      \|\partial\mathcal{L}/\partial\theta_i^{(m)}\|_2$
      for all $p_i \in \mathcal{S}$
    \State $r_m \gets \max(\epsilon,
      \sqrt{|\mathcal{S}|^{-1}\sum_i(g_i^{(m)})^2})$
\EndFor
\State $s_i \gets \sum_m g_i^{(m)}/r_m$
\State $\mathcal{G}_{L-1} \gets
  \{p_i\in\mathcal{S}:s_i\geq\tau\}$
\State $n_{\max} \gets |\mathcal{G}_{L-1}|$ and compute
  $\{n_l\}_{l=0}^{L-1}$ using Eq.~\eqref{eq:progression}
\For{$l=L-1$ \textbf{down to} $1$}
    \State Freeze $\Theta$ and accumulate gradients of $\mathcal{L}$
      over all views in $\mathcal{D}$ on $\mathcal{G}_l$
    \State Recompute normalized $s_i$ on $\mathcal{G}_l$ as above
    \State $\mathcal{E}_l \gets$ the $n_l-n_{l-1}$ primitives in
      $\mathcal{G}_l$ with lowest $s_i$
    \State $\mathcal{G}_{l-1} \gets \mathcal{G}_l \setminus
      \mathcal{E}_l$
\EndFor
\State \Return $\mathcal{G}_0,\{\mathcal{E}_l\}_{l=1}^{L-1}$
\end{algorithmic}
\end{algorithm}

\begin{algorithm}[h]
\caption{Quantization-aware multi-level fine-tuning}
\label{alg:qat}
\begin{algorithmic}
\Require Hierarchy $\{\mathcal{G}_l\}_{l=0}^{L-1}$, training views
  $\mathcal{D}$, iterations $T$, sparsity weight $\lambda$
\Ensure Fine-tuned quantization-ready parameters $\Theta$
\State Enable simulated quantization $Q(\cdot)$ with straight-through
  estimator for all quantized parameter groups
\For{$t=1$ \textbf{to} $T$}
    \State Sample level $l \sim \mathrm{Uniform}\{0,\dots,L-1\}$
    \State Sample training view $(c,I_c)\sim\mathcal{D}$
    \State Render $\hat{I}_c =
      \mathcal{R}(c,\{Q(\theta_i):p_i\in\mathcal{G}_l\})$
    \State Minimize $\mathcal{L}(\hat{I}_c, I_c) +
      \lambda\sum_{p_i\in\mathcal{G}_l}\|\theta_i^{(a)}\|_1$
    \State Update $\Theta$ via backprop with
      $\partial Q/\partial\theta \approx I$
\EndFor
\State Calibrate quantization ranges on the fine-tuned model
\State \Return $\Theta$
\end{algorithmic}
\end{algorithm}

\begin{algorithm}[h]
\caption{Progressive coding and decoding}
\label{alg:coding}
\begin{algorithmic}
\Require Fine-tuned hierarchy $\mathcal{G}_0,
  \{\mathcal{E}_l\}_{l=1}^{L-1}$, quantized parameters $\Theta$
\Ensure Progressive chunks $\{z_l\}_{l=0}^{L-1}$
\State Train one geometry dictionary $D_g$ and one appearance
  dictionary $D_a$ on the serialized fine-tuned model
\State $z_0 \gets \mathrm{Zstd}_{D_g, D_a}(
  \mathrm{Serialize}(Q(\mathcal{G}_0)))$
\For{$l=1$ \textbf{to} $L-1$}
    \State $z_l \gets \mathrm{Zstd}_{D_g, D_a}(
      \mathrm{Serialize}(Q(\mathcal{E}_l)))$
\EndFor
\State \textbf{Decode} level $l$: decompress and concatenate
  $\{z_0,\dots,z_l\}$
\State \Return $\{z_l\}_{l=0}^{L-1}$
\end{algorithmic}
\end{algorithm}

\section{Additional Ablations}
\label{supp:additional_ablations}

This section provides additional analyses of the hierarchy construction
and multi-level optimization. The extended ranking experiment uses a
wider diagnostic rate range, while the number-of-levels, lower-bound,
adaptive upper-bound, PUP-3DGS, convergence, and seed studies follow the
main experimental protocol.

\subsection{Score Normalization}
\label{supp:score_normalization}

We ablate the normalization in Eq.~\eqref{eq:importance} by comparing
the proposed score with a raw sum of the gradient norms across parameter
groups. We also include two single-attribute criteria based only on
positional gradients or opacity. All variants use the same pretrained
checkpoints, target primitive counts, hierarchy construction, and
fine-tuning protocol.

\begin{figure}[t]
    \centering
    \includegraphics[width=0.95\linewidth]{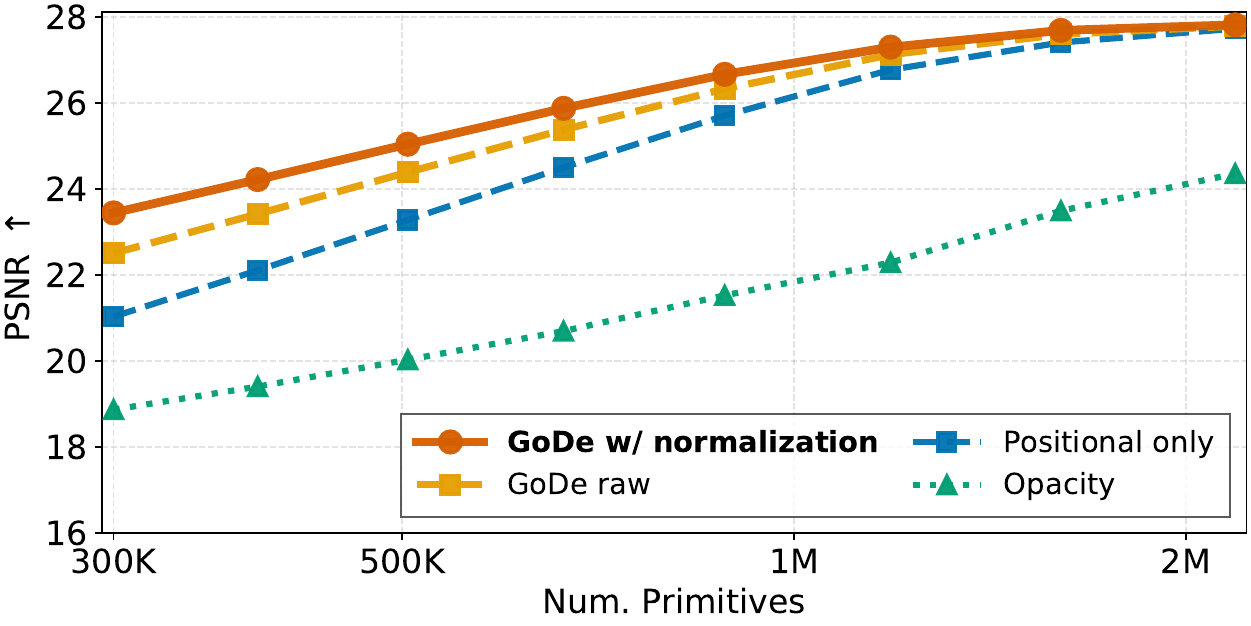}
    \caption{\textbf{Score-normalization ablation on Mip-NeRF360.}
    PSNR at matched active primitive counts for the proposed normalized
    multi-parameter score, its unnormalized counterpart, and
    single-attribute criteria.}
    \label{fig:score_normalization_ablation}
\end{figure}

Fig.~\ref{fig:score_normalization_ablation} shows that normalizing the
parameter groups consistently improves the ranking, with the largest
gains at aggressive pruning rates. Using all parameter groups also
substantially outperforms positional-gradient or opacity-only ranking,
supporting both the normalization and the multi-parameter aggregation
in Eq.~\eqref{eq:importance}.

\subsection{Number of Levels}
\label{supp:number_of_levels}

\begin{figure}[t]
    \centering
    \includegraphics[width=\linewidth]{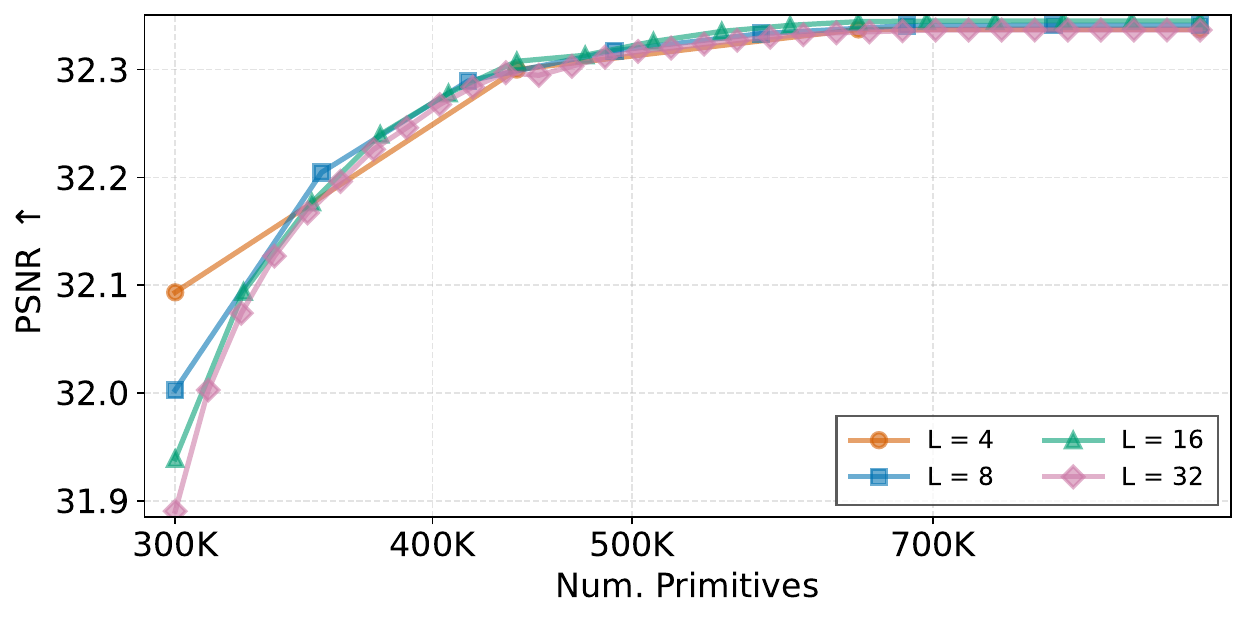}
    \caption{\textbf{Number of levels.} PSNR versus active primitive
    count on the Mip-NeRF360 \textit{bonsai} scene for
    $L\in\{4,8,16,32\}$.}
    \label{fig:number_of_levels}
\end{figure}

Fig.~\ref{fig:number_of_levels} shows that the curves are nearly
identical across most of the rate range. Increasing $L$ mainly provides
finer granularity between the fixed hierarchy endpoints, with only
minor differences at the lowest operating points. We use $L{=}8$ by
default as a practical trade-off between rate granularity and the
number of decoded levels. When finer granularity is required,
$L{=}16$ or $L{=}32$ remain viable with minimal impact on overall
reconstruction quality.

\subsection{Lower Hierarchy Bound}
\label{supp:lower_hierarchy_bound}

\begin{figure}[t]
    \centering
    \includegraphics[width=\linewidth]{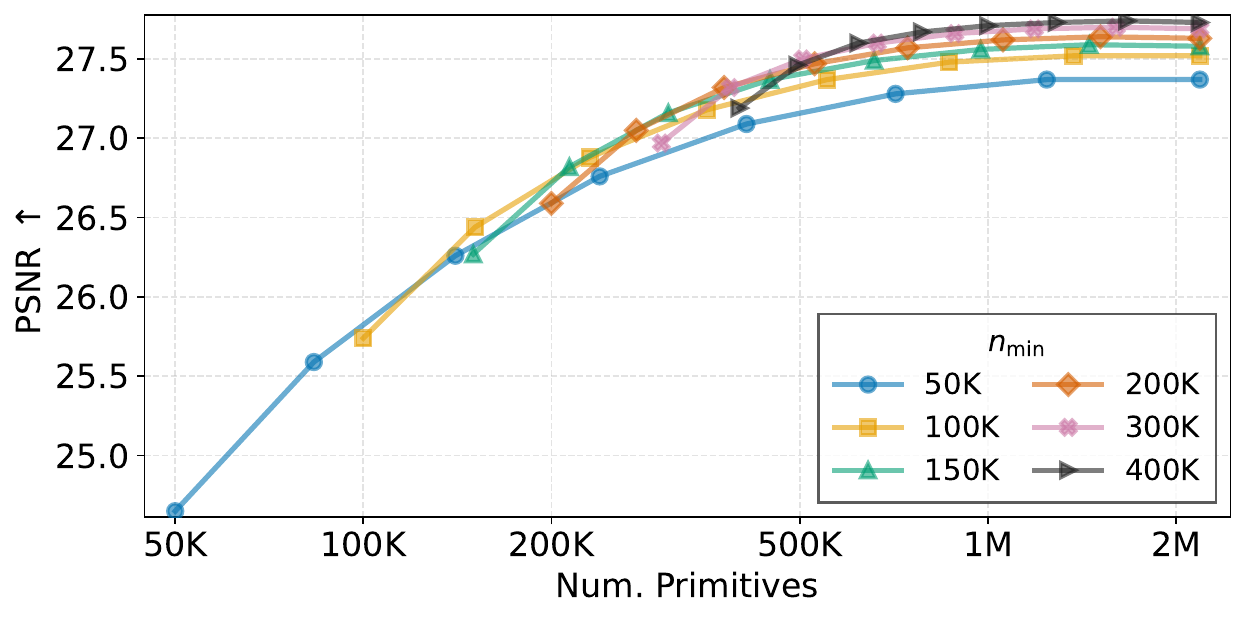}
    \caption{\textbf{Lower hierarchy bound.} PSNR versus active
    primitive count on Mip-NeRF360 for different values of $n_{\min}$.}
    \label{fig:lower_hierarchy_bound}
\end{figure}

Fig.~\ref{fig:lower_hierarchy_bound} studies the choice of the lowest
supported primitive count $n_{\min}$. Smaller values extend the
progressive curve toward lower-complexity models, at the cost of a more
aggressive hierarchy and some degradation in reconstruction quality.
We use an absolute rather than relative lower bound to directly control
the minimum model complexity. Thus, $n_{\min}$ is a
deployment parameter rather than a quantity with a universal optimum, controlling the supported complexity range and associated storage and
rendering cost.

For evaluation, we fix $n_{\min}$ per dataset and representation family,
without scene-specific tuning, and choose it so that the lowest
operating points cover a similar rate range to PCGS. We use $300$k
explicit primitives and $100$k anchors on Mip-NeRF360, and $100$k
primitives or anchors on Tanks\&Temples and DeepBlending. Notably, $300$k primitives correspond to roughly $10\%$ of the
original Mip-NeRF360 models, which contain about $3$M Gaussians on
average, so this setting already covers a highly compressed regime.

\subsection{Adaptive Upper Bound}
\label{supp:adaptive_upper_bound}

Unlike the lower bound, which specifies a deployment-dependent
quality--complexity trade-off, the upper bound can be selected adaptively
by removing inactive capacity without measurable quality degradation.
A fixed upper endpoint can conservatively discard a predetermined
fraction of the pretrained model; in our initial setting, retaining
$75\%$ of the primitives preserved essentially full-model quality.
However, the normalized importance scores reveal that the amount of
inactive capacity varies substantially across scenes. We therefore use
a scene-adaptive upper bound that removes primitives with score below
$10^{-12}$ before constructing the hierarchy. Across the nine Mip-NeRF360 scenes, this rule retains on average
$62.86\%$ of the pretrained Gaussians, corresponding to $37.14\%$
pruning, while preserving PSNR, SSIM, and LPIPS at the reported
precision. Compared with the conservative $75\%$ retain setting, the
adaptive rule therefore removes more inactive capacity
while preserving full-model reconstruction quality.

\begin{table*}[t]
\centering
\caption{\textbf{Adaptive upper-bound analysis on Mip-NeRF360.}
Dataset-averaged reconstruction quality after retaining a fixed fraction
of the pretrained Gaussians, compared with the proposed scene-adaptive
criterion. The adaptive rule retains $62.864\%$ of primitives on average
while preserving full-model quality at the reported precision.}
\label{tab:retain_percent}
\scriptsize
\setlength{\tabcolsep}{4pt}
\renewcommand{\arraystretch}{1.05}
\resizebox{0.98\textwidth}{!}{
\begin{tabular}{l cccc cccccccccc c}
\toprule
Retained
& 1.00 & 0.90 & 0.80 & \cellcolor{black!10}0.75 & 0.65 & 0.60 & 0.55 & 0.50 & 0.45 & 0.40 & 0.35 & 0.30 & 0.25 & 0.20
& \cellcolor{black!10}\makecell{0.63\\\textit{(Adaptive)}} \\
\midrule
PSNR $\uparrow$
& 27.82 & 27.82 & 27.82 & \cellcolor{black!10}27.82 & 27.81 & 27.79 & 27.72 & 27.59 & 27.37 & 27.03 & 26.53 & 25.85 & 24.97 & 23.83
& \cellcolor{black!10}27.82 \\
SSIM $\uparrow$
& 0.830 & 0.830 & 0.830 & \cellcolor{black!10}0.830 & 0.830 & 0.829 & 0.827 & 0.824 & 0.819 & 0.812 & 0.802 & 0.789 & 0.771 & 0.746
& \cellcolor{black!10}0.830 \\
LPIPS $\downarrow$
& 0.239 & 0.239 & 0.239 & \cellcolor{black!10}0.240 & 0.240 & 0.242 & 0.244 & 0.248 & 0.254 & 0.261 & 0.272 & 0.285 & 0.304 & 0.329
& \cellcolor{black!10}0.239 \\
\bottomrule
\end{tabular}
}
\end{table*}

\subsection{Importance Ranking under the PUP-3DGS Protocol}
\label{supp:pup_protocol}


\begin{table}[htbp]
\caption{\textbf{Comparison under the PUP-3DGS protocol.}
Both methods use the same pretrained checkpoints, $90\%$ total pruning,
and a $10$k-iteration optimization budget.}
\label{tab:vs_pup}
\centering
\footnotesize
\setlength{\tabcolsep}{3.5pt}
\renewcommand{\arraystretch}{1.08}
\resizebox{0.98\columnwidth}{!}{%
\begin{tabular}{l c cc ccc}
\toprule
Dataset & Method
& \shortstack{Wall-clock\\time (s)}
& \shortstack{Ranking\\time (s)}
& PSNR$\uparrow$ & SSIM$\uparrow$ & LPIPS$\downarrow$ \\
\midrule

Mip-NeRF360
& PUP
& 361 & 16
& 26.55 & 0.784 & 0.271 \\

& GoDe
& \textbf{315} & \textbf{9}
& \textbf{26.88} & \textbf{0.795} & \textbf{0.263} \\

\midrule

Tanks\&Temples
& PUP
& 215 & 17
& 22.45 & 0.797 & 0.251 \\

& GoDe
& \textbf{183} & \textbf{6}
& \textbf{22.88} & \textbf{0.804} & \textbf{0.247} \\

\midrule

DeepBlending
& PUP
& 313 & 15
& 29.57 & \textbf{0.904} & \textbf{0.260} \\

& GoDe
& \textbf{279} & \textbf{9}
& \textbf{29.58} & 0.902 & 0.262 \\

\bottomrule
\end{tabular}%
}
\end{table}

Table~\ref{tab:vs_pup} evaluates our importance criterion under the
original PUP-3DGS~\cite{hanson2025pup} pruning protocol. Both methods
start from the same pretrained 3DGS checkpoint, prune $80\%$ of the
primitives, fine-tune for $5$k iterations, prune $50\%$ of the
remainder, and fine-tune for another $5$k. The final primitive count,
loss, schedule, and total $10$k-iteration optimization budget are
identical; only the importance criterion changes. Our score improves
PSNR by $0.33$\,dB on Mip-NeRF360 and $0.43$\,dB on
Tanks\&Temples, while also reducing ranking and total wall-clock time.
On DeepBlending the two criteria are effectively tied in PSNR, with
PUP-3DGS retaining slightly better SSIM and LPIPS.

\subsection{Convergence and Seed Robustness}
\label{supp:convergence_robustness}

\begin{figure}[t]
    \centering
    \includegraphics[width=\linewidth]{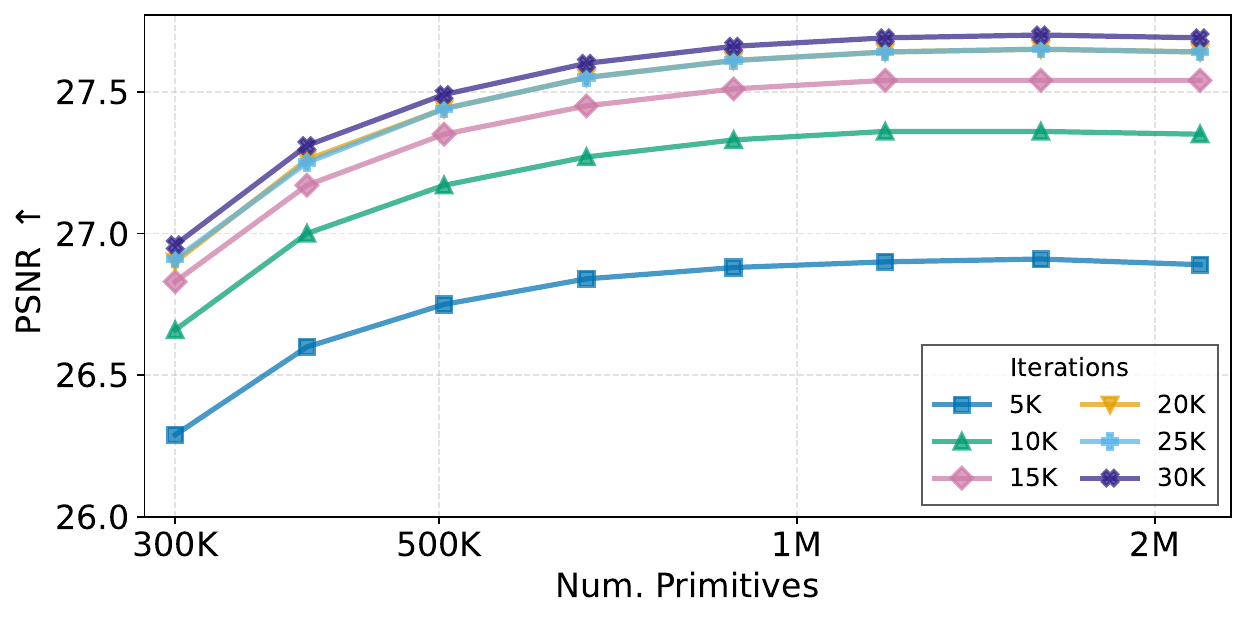}
    \caption{\textbf{Multi-level convergence on Mip-NeRF360.}
    PSNR across the eight decoded prefixes at different optimization
    iterations, averaged over the nine scenes.}
    \label{fig:convergence}
\end{figure}

Fig.~\ref{fig:convergence} reports per-LoD convergence averaged over
the nine Mip-NeRF360 scenes. Quality is already close to its final
value after $20$k iterations, with only minor changes through $30$k
iterations across all levels. We therefore keep the standard $30$k
schedule for consistency with common 3DGS optimization protocols; since
GoDe produces all LoDs in a single run, the additional iterations add
only a modest one-time training cost.

Five independent runs provide the complementary seed analysis in
Table~\ref{tab:avg_5runs}. Across levels, the PSNR standard deviation
is at most $0.016$\,dB, while SSIM and LPIPS deviations remain on the
order of $10^{-4}$.

\begin{table}[t]
\centering
\caption{\textbf{Seed robustness on Mip-NeRF360.}
Mean and standard deviation over five independent runs of GoDe (3DGS)
across all eight decoded levels.}
\label{tab:avg_5runs}
\footnotesize
\setlength{\tabcolsep}{8pt}
\renewcommand{\arraystretch}{1.05}
\resizebox{0.98\columnwidth}{!}{
\begin{tabular}{c ccc}
\toprule
Rate & PSNR $\uparrow$ & SSIM $\uparrow$ & LPIPS $\downarrow$ \\
\midrule
0 & $26.963 \pm 0.015$ & $0.7812 \pm 0.0003$ & $0.3219 \pm 0.0002$ \\
1 & $27.329 \pm 0.016$ & $0.7969 \pm 0.0002$ & $0.3021 \pm 0.0002$ \\
2 & $27.493 \pm 0.014$ & $0.8071 \pm 0.0001$ & $0.2858 \pm 0.0003$ \\
3 & $27.606 \pm 0.008$ & $0.8129 \pm 0.0001$ & $0.2742 \pm 0.0003$ \\
4 & $27.653 \pm 0.011$ & $0.8171 \pm 0.0002$ & $0.2650 \pm 0.0001$ \\
5 & $27.697 \pm 0.010$ & $0.8188 \pm 0.0003$ & $0.2581 \pm 0.0001$ \\
6 & $27.691 \pm 0.014$ & $0.8202 \pm 0.0003$ & $0.2539 \pm 0.0001$ \\
7 & $27.698 \pm 0.014$ & $0.8198 \pm 0.0003$ & $0.2522 \pm 0.0002$ \\
\bottomrule
\end{tabular}
}
\end{table}

\section{Cross-Backbone Transfer}
\label{supp:backbone_controls}

\begin{figure*}[t]
    \centering
    \includegraphics[width=0.8\linewidth]{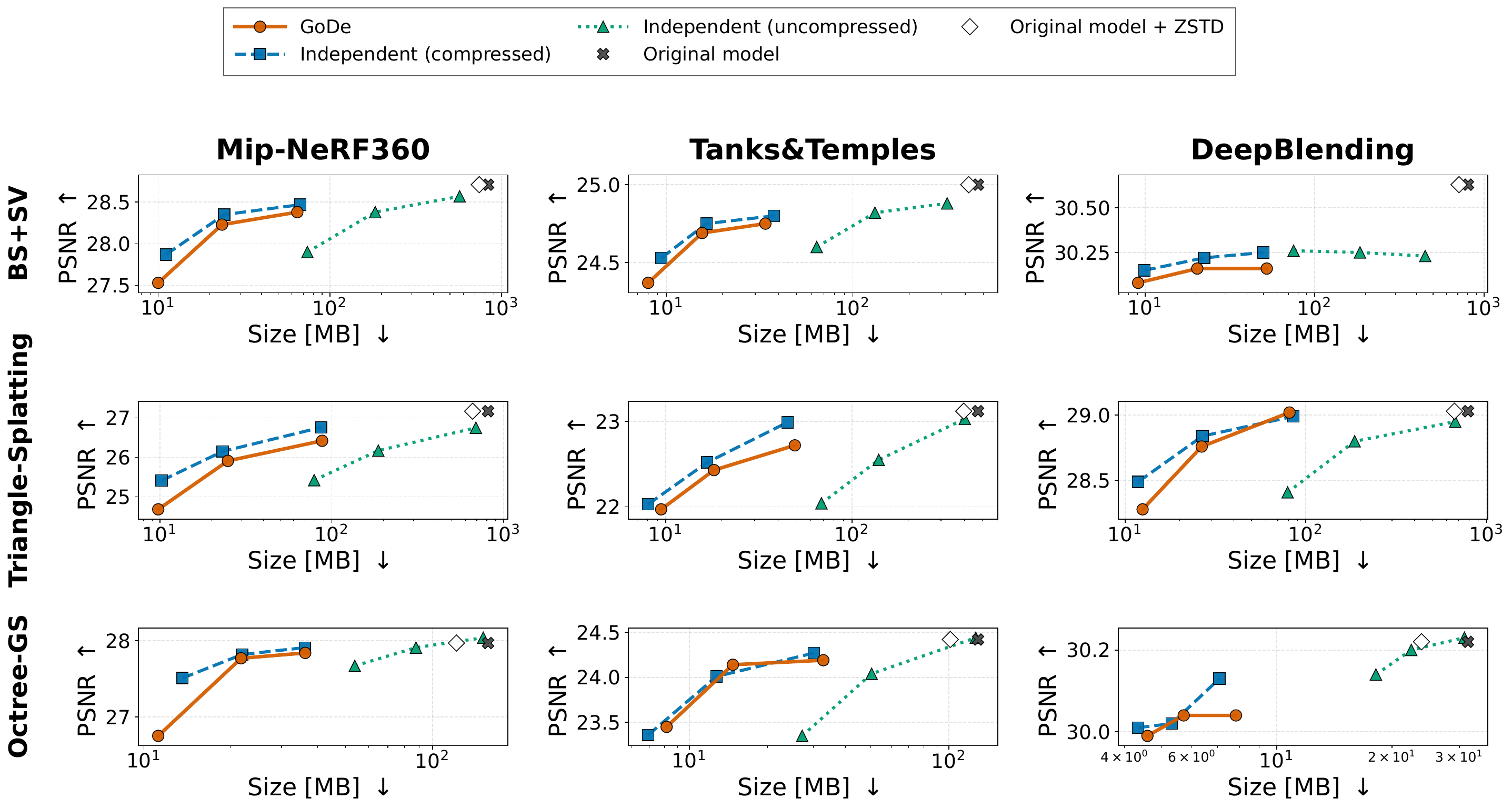}
    \caption{\textbf{Cross-backbone transfer across datasets.}
    Rate--distortion comparison of GoDe with independently optimized
    same-backbone models for BS+SV, Triangle Splatting, and Octree-GS
    on Mip-NeRF360, Tanks\&Temples, and DeepBlending. Independent
    models are shown before and after the same quantization and
    compression pipeline used by GoDe; the original model is included
    with and without lossless Zstandard compression for reference.}
    \label{fig:transferability_supp}
\end{figure*}

Fig.~\ref{fig:transferability_supp} extends the transferability analysis
to BS+SV, Triangle Splatting, and Octree-GS across all three datasets,
for which no established same-representation compression baselines are
available. For each backbone, we select three representative GoDe
levels (R2, R4, and R7) and train an independent model with the same
primitive or anchor count for $30$k iterations. Thus, the comparison is
matched in representation complexity rather than compressed file size;
small size differences can remain after coding. Each reference is
evaluated both uncompressed and after applying the same quantization
and compression pipeline used by GoDe.

We limit the comparison to three operating points because each point
requires a full independent training run for every scene, backbone, and
dataset. Despite this substantially more expensive rate-specific
optimization, GoDe remains close to the independent models while
providing all eight levels from a single progressive representation.
The rate--distortion plots are shown in
Fig.~\ref{fig:transferability_supp}, while
Tables~\ref{tab:baseline_bssv}--\ref{tab:baseline_octree} report the
complete PSNR, SSIM, LPIPS, and size results. 

\begin{table*}[tp]
\centering
\caption{\textbf{Cross-backbone transfer with BS+SV.} Independent reference models are separately trained for $30$k iterations at the primitive counts corresponding to GoDe levels R2, R4, and R7, and are reported before and after applying the same quantization and compression pipeline used by GoDe.}
\label{tab:baseline_bssv}
\scriptsize
\setlength{\tabcolsep}{2.35pt}
\renewcommand{\arraystretch}{0.98}
\resizebox{0.9\textwidth}{!}{%
\begin{tabular}{l c | *{4}{c} | *{4}{c} | *{4}{c}}
\toprule
\multirow{2}{*}{Dataset} & \multirow{2}{*}{Rate} &
\multicolumn{4}{c}{\textbf{GoDe (BS+SV)}} &
\multicolumn{4}{c}{Independent training w/ compression} &
\multicolumn{4}{c}{Independent training w/o compression} \\
\cmidrule(lr){3-6}
\cmidrule(lr){7-10}
\cmidrule(l){11-14}
& &
PSNR$\uparrow$ & SSIM$\uparrow$ & LPIPS$\downarrow$ & Size [MB]$\downarrow$ &
PSNR$\uparrow$ & SSIM$\uparrow$ & LPIPS$\downarrow$ & Size [MB]$\downarrow$ &
PSNR$\uparrow$ & SSIM$\uparrow$ & LPIPS$\downarrow$ & Size [MB]$\downarrow$ \\
\midrule

\multirow{3}{*}{Mip-NeRF360}
 & 2 & 27.53 & 0.785 & 0.323 & 10.01  & 27.87 & 0.798 & 0.310 & 11.11  & 27.90 & 0.799 & 0.307 & 73.97 \\
 & 4 & 28.23 & 0.818 & 0.269 & 23.54  & 28.35 & 0.821 & 0.259 & 24.26  & 28.38 & 0.822 & 0.257 & 183.58 \\
 & 7 & 28.38 & 0.826 & 0.239 & 64.63  & 28.47 & 0.827 & 0.231 & 67.32  & 28.57 & 0.829 & 0.227 & 569.66 \\

\methodrule

\multirow{3}{*}{Tanks\&Temples}
 & 2 & 24.37 & 0.841 & 0.240 & 8.00   & 24.53 & 0.847 & 0.230 & 9.42   & 24.60 & 0.849 & 0.228 & 64.03 \\
 & 4 & 24.69 & 0.856 & 0.204 & 15.57  & 24.75 & 0.859 & 0.198 & 16.42  & 24.82 & 0.862 & 0.193 & 131.43 \\
 & 7 & 24.75 & 0.861 & 0.188 & 33.99  & 24.80 & 0.863 & 0.180 & 37.74  & 24.88 & 0.864 & 0.175 & 320.43 \\

\methodrule

\multirow{3}{*}{DeepBlending}
 & 2 & 30.08 & 0.902 & 0.344 & 9.10   & 30.15 & 0.903 & 0.341 & 9.91   & 30.26 & 0.906 & 0.334 & 75.19 \\
 & 4 & 30.16 & 0.906 & 0.324 & 20.32  & 30.22 & 0.907 & 0.320 & 22.29  & 30.25 & 0.907 & 0.315 & 185.32 \\
 & 7 & 30.16 & 0.907 & 0.315 & 52.23  & 30.25 & 0.907 & 0.305 & 49.92  & 30.23 & 0.907 & 0.301 & 449.52 \\

\bottomrule
\end{tabular}%
}
\end{table*}

\begin{table*}[tp]
\centering
\caption{\textbf{Cross-backbone transfer with Triangle Splatting.} Independent reference models are separately trained for $30$k iterations at the primitive counts corresponding to GoDe levels R2, R4, and R7, and are reported before and after applying the same quantization and compression pipeline used by GoDe.}
\label{tab:baseline_triangle}
\scriptsize
\setlength{\tabcolsep}{2.35pt}
\renewcommand{\arraystretch}{0.98}
\resizebox{0.9\textwidth}{!}{%
\begin{tabular}{l c | *{4}{c} | *{4}{c} | *{4}{c}}
\toprule
\multirow{2}{*}{Dataset} & \multirow{2}{*}{Rate} &
\multicolumn{4}{c}{\textbf{GoDe (Triangle-Splatting)}} &
\multicolumn{4}{c}{Independent training w/ compression} &
\multicolumn{4}{c}{Independent training w/o compression} \\
\cmidrule(lr){3-6}
\cmidrule(lr){7-10}
\cmidrule(l){11-14}
& &
PSNR$\uparrow$ & SSIM$\uparrow$ & LPIPS$\downarrow$ & Size [MB]$\downarrow$ &
PSNR$\uparrow$ & SSIM$\uparrow$ & LPIPS$\downarrow$ & Size [MB]$\downarrow$ &
PSNR$\uparrow$ & SSIM$\uparrow$ & LPIPS$\downarrow$ & Size [MB]$\downarrow$ \\
\midrule

\multirow{3}{*}{Mip-NeRF360}
 & 2 & 24.68 & 0.685 & 0.425 & 9.75   & 25.41 & 0.718 & 0.385 & 10.21  & 25.42 & 0.718 & 0.384 & 79.00 \\
 & 4 & 25.91 & 0.743 & 0.348 & 24.83  & 26.15 & 0.756 & 0.328 & 23.15  & 26.17 & 0.756 & 0.327 & 186.79 \\
 & 7 & 26.42 & 0.772 & 0.304 & 87.91  & 26.76 & 0.783 & 0.276 & 86.73  & 26.75 & 0.783 & 0.275 & 691.51 \\

\methodrule

\multirow{3}{*}{Tanks\&Temples}
 & 2 & 21.97 & 0.786 & 0.301 & 9.44   & 22.03 & 0.793 & 0.303 & 8.04   & 22.04 & 0.793 & 0.301 & 68.44 \\
 & 4 & 22.43 & 0.811 & 0.265 & 18.14  & 22.52 & 0.817 & 0.256 & 16.68  & 22.55 & 0.818 & 0.254 & 138.91 \\
 & 7 & 22.72 & 0.828 & 0.234 & 49.26  & 22.99 & 0.836 & 0.213 & 45.04  & 23.03 & 0.836 & 0.212 & 402.50 \\

\methodrule

\multirow{3}{*}{DeepBlending}
 & 2 & 28.28 & 0.873 & 0.400 & 12.48  & 28.49 & 0.881 & 0.384 & 11.79  & 28.41 & 0.880 & 0.383 & 79.29 \\
 & 4 & 28.76 & 0.883 & 0.372 & 26.52  & 28.84 & 0.887 & 0.356 & 26.80  & 28.80 & 0.887 & 0.355 & 186.29 \\
 & 7 & 29.02 & 0.889 & 0.349 & 81.15  & 28.99 & 0.887 & 0.329 & 85.23  & 28.95 & 0.885 & 0.329 & 671.23 \\

\bottomrule
\end{tabular}%
}
\end{table*}

\begin{table*}[tp]
\centering
\caption{\textbf{Cross-backbone transfer with Octree-GS.} Independent reference models are separately trained for $30$k iterations at the primitive counts corresponding to GoDe levels R2, R4, and R7, and are reported before and after applying the same quantization and compression pipeline used by GoDe.}
\label{tab:baseline_octree}
\scriptsize
\setlength{\tabcolsep}{2.35pt}
\renewcommand{\arraystretch}{0.98}
\resizebox{0.9\textwidth}{!}{%
\begin{tabular}{l c | *{4}{c} | *{4}{c} | *{4}{c}}
\toprule
\multirow{2}{*}{Dataset} & \multirow{2}{*}{Rate} &
\multicolumn{4}{c}{\textbf{GoDe (Octree-GS)}} &
\multicolumn{4}{c}{Independent training w/ compression} &
\multicolumn{4}{c}{Independent training w/o compression} \\
\cmidrule(lr){3-6}
\cmidrule(lr){7-10}
\cmidrule(l){11-14}
& &
PSNR$\uparrow$ & SSIM$\uparrow$ & LPIPS$\downarrow$ & Size [MB]$\downarrow$ &
PSNR$\uparrow$ & SSIM$\uparrow$ & LPIPS$\downarrow$ & Size [MB]$\downarrow$ &
PSNR$\uparrow$ & SSIM$\uparrow$ & LPIPS$\downarrow$ & Size [MB]$\downarrow$ \\
\midrule

\multirow{3}{*}{Mip-NeRF360}
 & 2 & 26.75 & 0.779 & 0.309 & 11.21  & 27.51 & 0.806 & 0.283 & 13.61  & 27.67 & 0.808 & 0.279 & 53.78 \\
 & 4 & 27.77 & 0.815 & 0.265 & 21.72  & 27.82 & 0.817 & 0.262 & 21.90  & 27.91 & 0.819 & 0.259 & 87.60 \\
 & 7 & 27.84 & 0.819 & 0.257 & 36.22  & 27.91 & 0.819 & 0.253 & 36.10  & 28.04 & 0.821 & 0.250 & 149.80 \\

\methodrule

\multirow{3}{*}{Tanks\&Temples}
 & 2 & 23.45 & 0.830 & 0.288 & 8.16   & 23.36 & 0.828 & 0.284 & 6.93   & 23.35 & 0.829 & 0.280 & 27.18 \\
 & 4 & 24.14 & 0.862 & 0.232 & 14.69  & 24.01 & 0.858 & 0.236 & 12.72  & 24.04 & 0.859 & 0.231 & 50.32 \\
 & 7 & 24.19 & 0.866 & 0.222 & 32.79  & 24.27 & 0.872 & 0.205 & 30.17  & 24.44 & 0.874 & 0.201 & 126.81 \\

\methodrule

\multirow{3}{*}{DeepBlending}
 & 2 & 29.99 & 0.899 & 0.350 & 4.59   & 30.02 & 0.901 & 0.347 & 4.34   & 30.14 & 0.901 & 0.344 & 18.16 \\
 & 4 & 30.04 & 0.900 & 0.346 & 5.71   & 30.02 & 0.900 & 0.342 & 5.31   & 30.20 & 0.902 & 0.339 & 22.47 \\
 & 7 & 30.04 & 0.900 & 0.344 & 7.82   & 30.13 & 0.903 & 0.337 & 7.08   & 30.23 & 0.903 & 0.335 & 30.94 \\

\bottomrule
\end{tabular}%
}
\end{table*}

\section{Quantization-Aware Fine-Tuning Details}
\label{supp:qat}

\noindent\textbf{General setup.}
QAT adapts the shared parameter vector $\Theta$ to perform well at
every decoded prefix simultaneously and under the quantization applied
at coding time.
Unless otherwise stated, we use $L{=}8$ output levels selected via
the geometric schedule of Eq.~\ref{eq:progression}. We set
$n_{\max}$ per scene by retaining primitives whose normalized
importance is at least $10^{-12}$. The lower hierarchy bound
$n_{\min}$ is fixed per dataset and representation family, without
scene-specific tuning, as detailed in
Sec.~\ref{supp:lower_hierarchy_bound}. All backbones are fine-tuned for
$30$k iterations.
Geometric attributes are frozen for the final $3$k iterations to
stabilize quantization ranges before calibration.

\noindent\textbf{Quantization scheme.}
We use two storage classes across all backbones.
Geometric attributes are stored in float16, while appearance attributes
are uniformly quantized to int8 using affine quantization:
\begin{equation}
s = \frac{x_{\max}-x_{\min}}{q_{\max}-q_{\min}}, \qquad
z = \mathrm{round}\!\left(q_{\min} - \frac{x_{\min}}{s}\right),
\end{equation}
\begin{equation}
x_q = \mathrm{clip}\!\left(
  \mathrm{round}\!\left(\tfrac{x}{s}+z\right),
  q_{\min}, q_{\max}\right), \qquad
\hat{x} = s(x_q - z).
\end{equation}
The forward pass uses $\hat{x}$ in place of $x$; the backward pass
uses the straight-through estimator $\partial Q/\partial x \approx I$.

\noindent\textbf{Per-backbone attribute mapping.}
For GoDe (3DGS) and GoDe (Triangle Splatting), appearance attributes
are the spherical harmonics coefficients.
For GoDe (BS+SV), they are the Spherical Voronoi site directions,
lobe colors, and temperatures.
For GoDe (Scaffold-GS) and GoDe (Octree-GS), they are the anchor
feature codes.
In all cases, opacity is treated as an appearance attribute and
quantized to int8; the remaining geometric attributes are stored in
float16.

\noindent\textbf{Fine-tuning objective.}
At each iteration, one level $l$ is sampled uniformly and the
fine-tuning objective is:
\begin{equation}
\min_{\Theta}\;
\mathcal{L}\!\left(
  \mathcal{R}\!\left(c,\{Q(\theta_i):p_i\in\mathcal{G}_l\}\right),
  I_c\right)
+ \lambda\sum_{p_i\in\mathcal{G}_l}\|\theta_i^{(a)}\|_1,
\end{equation}
where $\mathcal{L}$ is the backbone's standard photometric loss
(L1~+~D-SSIM) and the $\ell_1$ term encourages sparse appearance
parameters before coding.
We use $\lambda{=}0.05$ for all backbones and scenes.

\noindent\textbf{Optimizer and schedule.}
We follow each backbone's original optimizer settings (Adam with
backbone-specific learning rates), resetting the optimizer state at the
start of QAT. No learning-rate warmup is applied, and the backbone's
default decay schedule is reused for the QAT duration. Apart from the
attribute mapping and the fixed per-backbone settings reported above,
the same GoDe optimization procedure is used across all representations.

\section{Full Per-LoD Results}
\label{supp:fulltables}
\label{supp:qualitative}

\begin{table*}[t]
\centering
\caption{\textbf{Full per-LoD results of GoDe.} Dataset-averaged results for
all eight decoded levels and all five backbones. Size is cumulative and
includes all chunks required to decode the corresponding level.}
\label{tab:supp_full}
\scriptsize
\setlength{\tabcolsep}{3pt}
\renewcommand{\arraystretch}{1.0}
\resizebox{\textwidth}{!}{%
\begin{tabular}{lc ccccc ccccc ccccc}
\toprule
& & \multicolumn{5}{c}{Mip-NeRF360} & \multicolumn{5}{c}{Tanks\&Temples} & \multicolumn{5}{c}{DeepBlending} \\
\cmidrule(lr){3-7}\cmidrule(lr){8-12}\cmidrule(lr){13-17}
Backbone & LoD
& PSNR$\uparrow$ & SSIM$\uparrow$ & LPIPS$\downarrow$ & Size [MB]$\downarrow$ & FPS$\uparrow$
& PSNR$\uparrow$ & SSIM$\uparrow$ & LPIPS$\downarrow$ & Size [MB]$\downarrow$ & FPS$\uparrow$
& PSNR$\uparrow$ & SSIM$\uparrow$ & LPIPS$\downarrow$ & Size [MB]$\downarrow$ & FPS$\uparrow$ \\
\midrule

\multirow{8}{*}{Scaffold-GS}
 & 0 & 27.29 & 0.792 & 0.310 &  9.81 & 104 & 23.17 & 0.812 & 0.281 &  4.13 & 171 & 30.00 & 0.901 & 0.351 &  3.92 & 183 \\
 & 1 & 27.62 & 0.802 & 0.297 & 11.93 &  96 & 23.68 & 0.829 & 0.259 &  5.10 & 160 & 30.27 & 0.905 & 0.343 &  4.63 & 178 \\
 & 2 & 27.79 & 0.808 & 0.288 & 14.57 &  89 & 23.98 & 0.840 & 0.242 &  6.30 & 148 & 30.40 & 0.908 & 0.337 &  5.48 & 171 \\
 & 3 & 27.87 & 0.811 & 0.282 & 17.85 &  83 & 24.16 & 0.847 & 0.229 &  7.80 & 137 & 30.49 & 0.909 & 0.332 &  6.49 & 166 \\
 & 4 & 27.92 & 0.813 & 0.278 & 21.95 &  78 & 24.25 & 0.851 & 0.220 &  9.66 & 125 & 30.53 & 0.910 & 0.329 &  7.70 & 159 \\
 & 5 & 27.93 & 0.814 & 0.275 & 27.08 &  74 & 24.31 & 0.854 & 0.215 & 11.98 & 114 & 30.55 & 0.910 & 0.327 &  9.13 & 158 \\
 & 6 & 27.93 & 0.814 & 0.274 & 33.52 &  71 & 24.32 & 0.855 & 0.212 & 14.84 & 105 & 30.56 & 0.911 & 0.326 & 10.84 & 154 \\
 & 7 & 27.93 & 0.814 & 0.274 & 41.61 &  68 & 24.31 & 0.855 & 0.211 & 18.38 &  98 & 30.54 & 0.910 & 0.326 & 12.83 & 150 \\
\midrule

\multirow{8}{*}{Octree-GS}
 & 0 & 26.75 & 0.779 & 0.309 & 11.21 & 66 & 22.14 & 0.787 & 0.345 &  4.20 & 97 & 29.81 & 0.896 & 0.356 & 3.65 & 138 \\
 & 1 & 27.16 & 0.793 & 0.294 & 13.15 & 60 & 22.92 & 0.811 & 0.315 &  5.95 & 87 & 29.92 & 0.898 & 0.353 & 4.10 & 132 \\
 & 2 & 27.44 & 0.803 & 0.282 & 15.49 & 54 & 23.45 & 0.830 & 0.288 &  8.16 & 77 & 29.99 & 0.899 & 0.350 & 4.59 & 126 \\
 & 3 & 27.66 & 0.811 & 0.272 & 18.31 & 49 & 23.81 & 0.845 & 0.264 & 11.01 & 66 & 30.02 & 0.900 & 0.348 & 5.12 & 121 \\
 & 4 & 27.77 & 0.815 & 0.265 & 21.72 & 44 & 24.03 & 0.855 & 0.246 & 14.69 & 56 & 30.04 & 0.900 & 0.346 & 5.71 & 115 \\
 & 5 & 27.83 & 0.818 & 0.260 & 25.81 & 40 & 24.14 & 0.862 & 0.232 & 19.48 & 47 & 30.05 & 0.900 & 0.345 & 6.36 & 109 \\
 & 6 & 27.84 & 0.818 & 0.258 & 30.67 & 36 & 24.21 & 0.865 & 0.224 & 25.56 & 39 & 30.05 & 0.900 & 0.345 & 7.06 & 105 \\
 & 7 & 27.84 & 0.819 & 0.257 & 36.22 & 33 & 24.19 & 0.866 & 0.222 & 32.79 & 31 & 30.04 & 0.900 & 0.344 & 7.82 &  99 \\
\midrule

\multirow{8}{*}{BS+SV}
 & 0 & 27.53 & 0.785 & 0.323 &  9.97 & 213 & 23.62 & 0.814 & 0.285 &  4.74 & 357 & 29.70 & 0.892 & 0.370 &  4.80 & 365 \\
 & 1 & 27.91 & 0.801 & 0.300 & 12.74 & 192 & 24.08 & 0.830 & 0.260 &  6.15 & 328 & 29.96 & 0.899 & 0.355 &  6.61 & 323 \\
 & 2 & 28.11 & 0.811 & 0.283 & 16.32 & 171 & 24.37 & 0.841 & 0.240 &  8.00 & 301 & 30.08 & 0.902 & 0.344 &  9.10 & 289 \\
 & 3 & 28.23 & 0.818 & 0.269 & 20.96 & 152 & 24.54 & 0.849 & 0.224 & 10.43 & 274 & 30.13 & 0.904 & 0.336 & 12.52 & 260 \\
 & 4 & 28.33 & 0.823 & 0.252 & 30.70 & 124 & 24.69 & 0.856 & 0.204 & 15.57 & 236 & 30.16 & 0.906 & 0.324 & 20.32 & 218 \\
 & 5 & 28.37 & 0.825 & 0.245 & 39.73 & 108 & 24.74 & 0.859 & 0.195 & 20.37 & 208 & 30.17 & 0.907 & 0.319 & 28.06 & 192 \\
 & 6 & 28.38 & 0.826 & 0.241 & 51.34 &  92 & 24.75 & 0.860 & 0.190 & 26.51 & 183 & 30.17 & 0.907 & 0.316 & 38.57 & 163 \\
 & 7 & 28.38 & 0.826 & 0.239 & 65.33 &  80 & 24.75 & 0.861 & 0.188 & 33.99 & 162 & 30.16 & 0.907 & 0.315 & 52.23 & 136 \\
\midrule

\multirow{8}{*}{3DGS}
 & 0 & 26.97 & 0.781 & 0.322 & 11.31 & 270 & 22.49 & 0.783 & 0.325 &  4.12 & 485 & 29.01 & 0.885 & 0.378 &  4.25 & 562 \\
 & 1 & 27.32 & 0.797 & 0.302 & 14.29 & 245 & 23.12 & 0.804 & 0.297 &  5.44 & 421 & 29.41 & 0.893 & 0.362 &  5.94 & 486 \\
 & 2 & 27.50 & 0.807 & 0.286 & 18.16 & 220 & 23.51 & 0.819 & 0.274 &  7.23 & 374 & 29.58 & 0.897 & 0.350 &  8.37 & 429 \\
 & 3 & 27.60 & 0.813 & 0.274 & 23.21 & 198 & 23.79 & 0.829 & 0.254 &  9.71 & 324 & 29.63 & 0.899 & 0.340 & 11.92 & 368 \\
 & 4 & 27.66 & 0.817 & 0.265 & 29.77 & 176 & 23.96 & 0.837 & 0.238 & 13.17 & 277 & 29.67 & 0.900 & 0.333 & 17.11 & 319 \\
 & 5 & 27.69 & 0.819 & 0.258 & 38.19 & 156 & 24.05 & 0.841 & 0.227 & 17.91 & 236 & 29.67 & 0.901 & 0.328 & 24.69 & 270 \\
 & 6 & 27.70 & 0.820 & 0.254 & 48.74 & 137 & 24.09 & 0.843 & 0.221 & 24.37 & 195 & 29.67 & 0.901 & 0.325 & 35.49 & 228 \\
 & 7 & 27.69 & 0.820 & 0.252 & 61.45 & 122 & 24.09 & 0.844 & 0.218 & 33.11 & 172 & 29.65 & 0.900 & 0.325 & 50.55 & 191 \\
\midrule

\multirow{8}{*}{Triangle Splatting}
 & 0 & 24.68 & 0.685 & 0.460 &  9.75 & 203 & 20.87 & 0.738 & 0.357 &  5.00 & 371 & 26.87 & 0.847 & 0.446 &  5.94 & 259 \\
 & 1 & 25.30 & 0.711 & 0.425 & 13.31 & 180 & 21.59 & 0.767 & 0.325 &  6.87 & 337 & 27.77 & 0.863 & 0.420 &  8.61 & 237 \\
 & 2 & 25.65 & 0.729 & 0.397 & 18.16 & 157 & 21.97 & 0.786 & 0.301 &  9.44 & 297 & 28.28 & 0.873 & 0.400 & 12.48 & 214 \\
 & 3 & 25.91 & 0.743 & 0.371 & 24.83 & 133 & 22.24 & 0.800 & 0.281 & 13.06 & 256 & 28.57 & 0.879 & 0.385 & 18.19 & 191 \\
 & 4 & 26.09 & 0.754 & 0.348 & 34.07 & 111 & 22.43 & 0.811 & 0.265 & 18.14 & 214 & 28.76 & 0.883 & 0.372 & 26.52 & 162 \\
 & 5 & 26.25 & 0.763 & 0.328 & 46.85 &  90 & 22.58 & 0.820 & 0.251 & 25.30 & 172 & 28.88 & 0.886 & 0.362 & 38.70 & 132 \\
 & 6 & 26.36 & 0.770 & 0.312 & 64.39 &  72 & 22.67 & 0.826 & 0.239 & 35.30 & 132 & 28.96 & 0.888 & 0.354 & 56.07 & 102 \\
 & 7 & 26.42 & 0.772 & 0.304 & 87.91 &  57 & 22.72 & 0.828 & 0.234 & 49.26 & 101 & 29.02 & 0.889 & 0.349 & 81.15 &  77 \\

\bottomrule
\end{tabular}%
}
\end{table*}

\begin{figure*}[p]
    \centering
    \includegraphics[width=\linewidth]{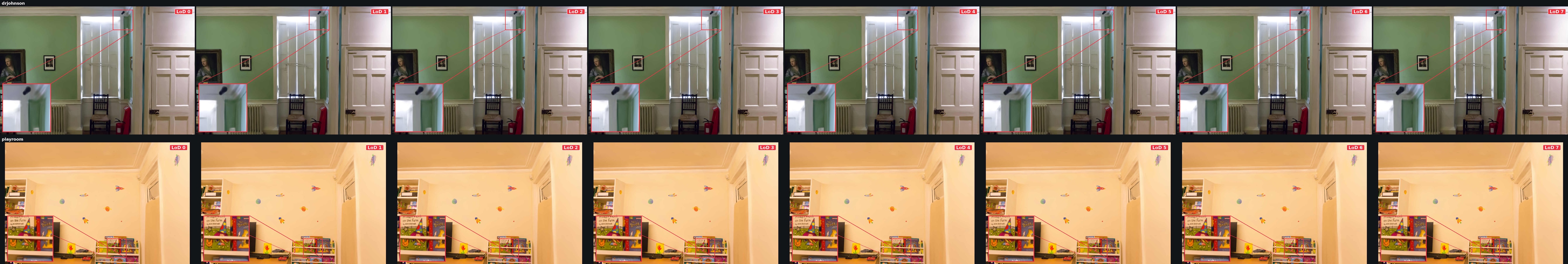}
    \caption{\textbf{Full LoD progression on DeepBlending.}
    All eight decoded levels using the 3DGS-MCMC backbone.}
    \label{fig:full_lod_db}

    \vspace{0.8em}

    \includegraphics[width=\linewidth]{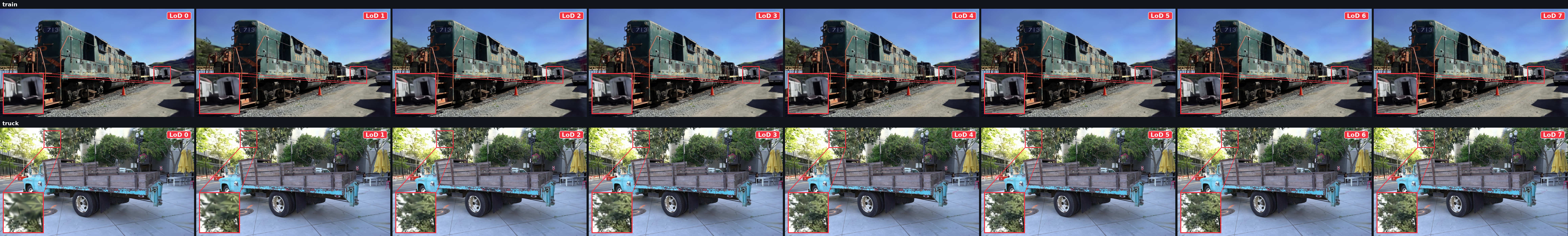}
    \caption{\textbf{Full LoD progression on Tanks\&Temples.}
    All eight decoded levels using the 3DGS-MCMC backbone.}
    \label{fig:full_lod_tat}

    \vspace{0.8em}

    \includegraphics[width=\linewidth]{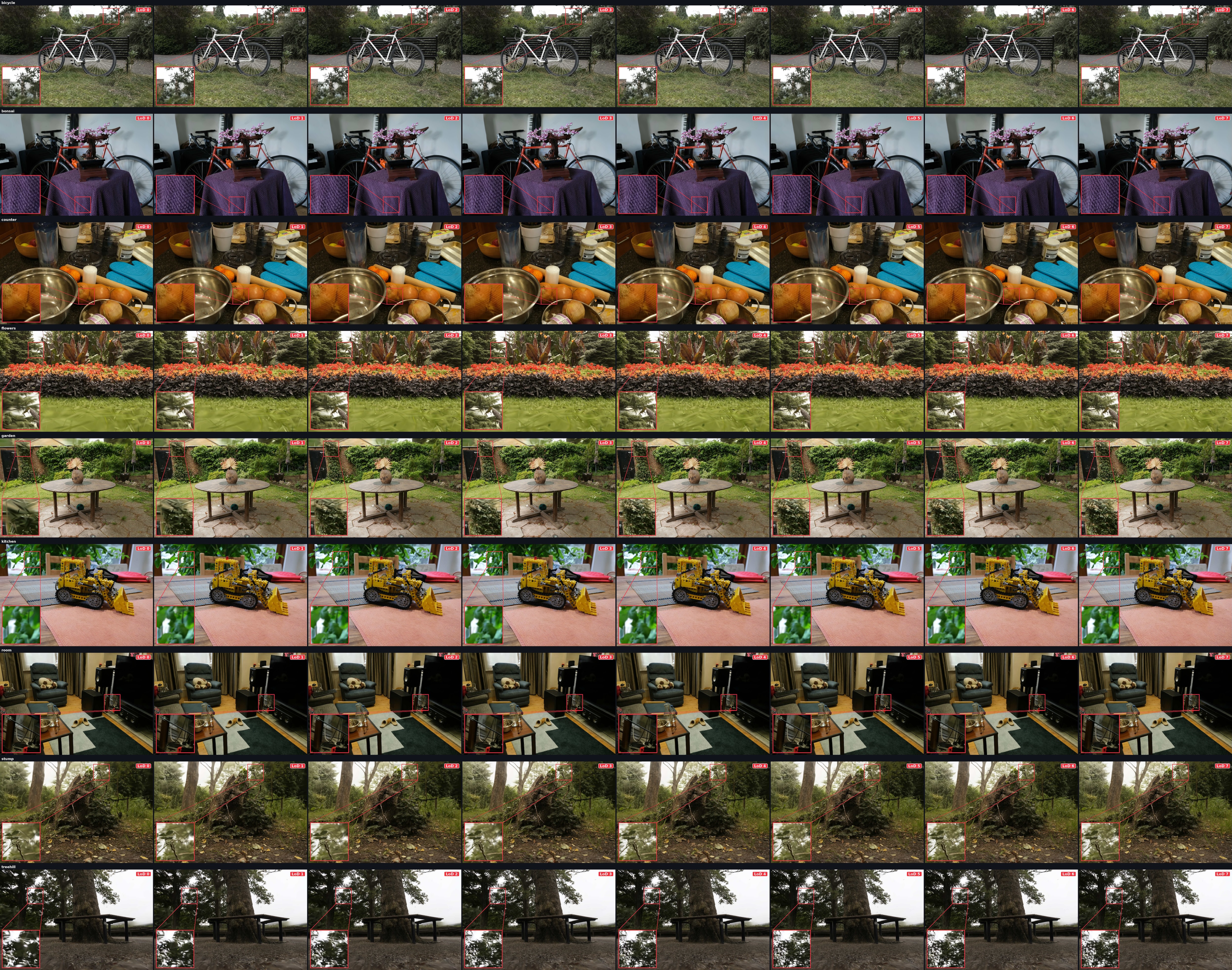}
    \caption{\textbf{Full LoD progression on Mip-NeRF360.}
    All eight decoded levels using the 3DGS-MCMC backbone.}
    \label{fig:full_lod_360}
\end{figure*}

Table~\ref{tab:supp_full} reports the complete dataset-averaged results
of GoDe for all eight decoded levels and all five backbones on
Mip-NeRF360, Tanks\&Temples, and DeepBlending. Size is cumulative and
includes all chunks required to decode the corresponding level.
Figs.~\ref{fig:full_lod_db}--\ref{fig:full_lod_360} show the
corresponding qualitative progression for representative scenes from
each benchmark, using the 3DGS-MCMC backbone. Each successive level adds
the primitives of one enhancement layer while preserving those already
present in coarser levels, as defined by Eq.~\eqref{eq:nesting}.

\end{document}